\documentclass[conference]{IEEEtran}
\IEEEoverridecommandlockouts

\usepackage{cite}
\usepackage{amsmath,amssymb,amsfonts}
\usepackage{algorithmic}
\usepackage{graphicx}
\usepackage{textcomp}
\usepackage{xcolor}

\usepackage{makecell}
\usepackage{multirow}
\usepackage{booktabs}
\usepackage{adjustbox}
\usepackage{hyperref}
\usepackage{cleveref}
\usepackage{eso-pic}

\def\BibTeX{{\rm B\kern-.05em{\sc i\kern-.025em b}\kern-.08em
    T\kern-.1667em\lower.7ex\hbox{E}\kern-.125emX}}
\begin{document}

\AddToShipoutPictureFG*{%
  \AtPageUpperLeft{%
    \raisebox{-14pt}[0pt][0pt]{%
      \makebox[\paperwidth]{%
        \parbox[t]{0.80\paperwidth}{\footnotesize\noindent
          \textcopyright{} 2026 IEEE. Personal use of this material is permitted.
          Permission from IEEE must be obtained for all other uses, in any current
          or future media, including reprinting/republishing this material for
          advertising or promotional purposes, creating new collective works, for
          resale or redistribution to servers or lists, or reuse of any copyrighted
          component of this work in other works.%
        }%
      }%
    }%
  }%
}

\title{Occlusion-Aware Panoptic Segmentation with Joint Position Embedding and Occlusion-Level Attention}

\author{\IEEEauthorblockN{Wenbo Wei, Jun Wang, Shan Raza, and Abhir Bhalerao}
\IEEEauthorblockA{Department of Computer Science, University of Warwick, Coventry, UK}
\IEEEauthorblockA{\{wenbo.wei, jun.wang.3, shan.raza, abhir.bhalerao\}@warwick.ac.uk}
}

\maketitle

\begin{abstract}
Panoptic segmentation in complex scenes remains challenging because of occlusions, yet modern approaches often neglect occlusion modelling. In this paper, we propose \textbf{P}osition \textbf{E}mbedding \textbf{M}odulation with \textbf{O}cclusion-\textbf{L}evel \textbf{A}ttention (PEMOLA), a novel occlusion-aware module that can be seamlessly integrated into transformer-based panoptic segmentation. To obtain occlusion cues, we train an occlusion classifier on the COCO-OLAC dataset. The classifier derives the occlusion-level attention, which serves as spatial guidance, while the occlusion labels are encoded into a learnable embedding to produce channel-wise weights. Through joint modulation, PEMOLA elegantly introduces the occlusion priors into the position embedding, thereby improving the occlusion modelling. We further annotate the Cityscapes dataset with occlusion levels, termed Cityscapes Occlusion Labels for All Computer Vision Tasks (Cityscapes-OLAC), following the same labelling protocol as COCO-OLAC, to evaluate the cross-dataset generalisation ability of PEMOLA. Extensive experiments on COCO-OLAC and Cityscapes-OLAC demonstrate that PEMOLA consistently improves panoptic segmentation quality while introducing minimal computational overhead. These results highlight the importance of occlusion modelling, where incorporating occlusion-level attention helps deliver robust panoptic segmentation under occlusion.\footnote{Code and dataset are available at \href{https://github.com/wenbo-wei/PEMOLA}{https://github.com/wenbo-wei/PEMOLA}.}
\end{abstract}

\begin{IEEEkeywords}
Panoptic Segmentation, Occlusion Modelling, Position Embedding
\end{IEEEkeywords}

\section{Introduction}
\label{sec:intro}

\begin{figure}[t]
    \centering
    \setlength{\tabcolsep}{3.6pt}
    \begin{tabular}{@{}cc@{}}
    \includegraphics[width=0.225\textwidth, height=2.6cm]{fig/intro/000000205514.pdf} & 
    \includegraphics[width=0.225\textwidth, height=2.6cm]{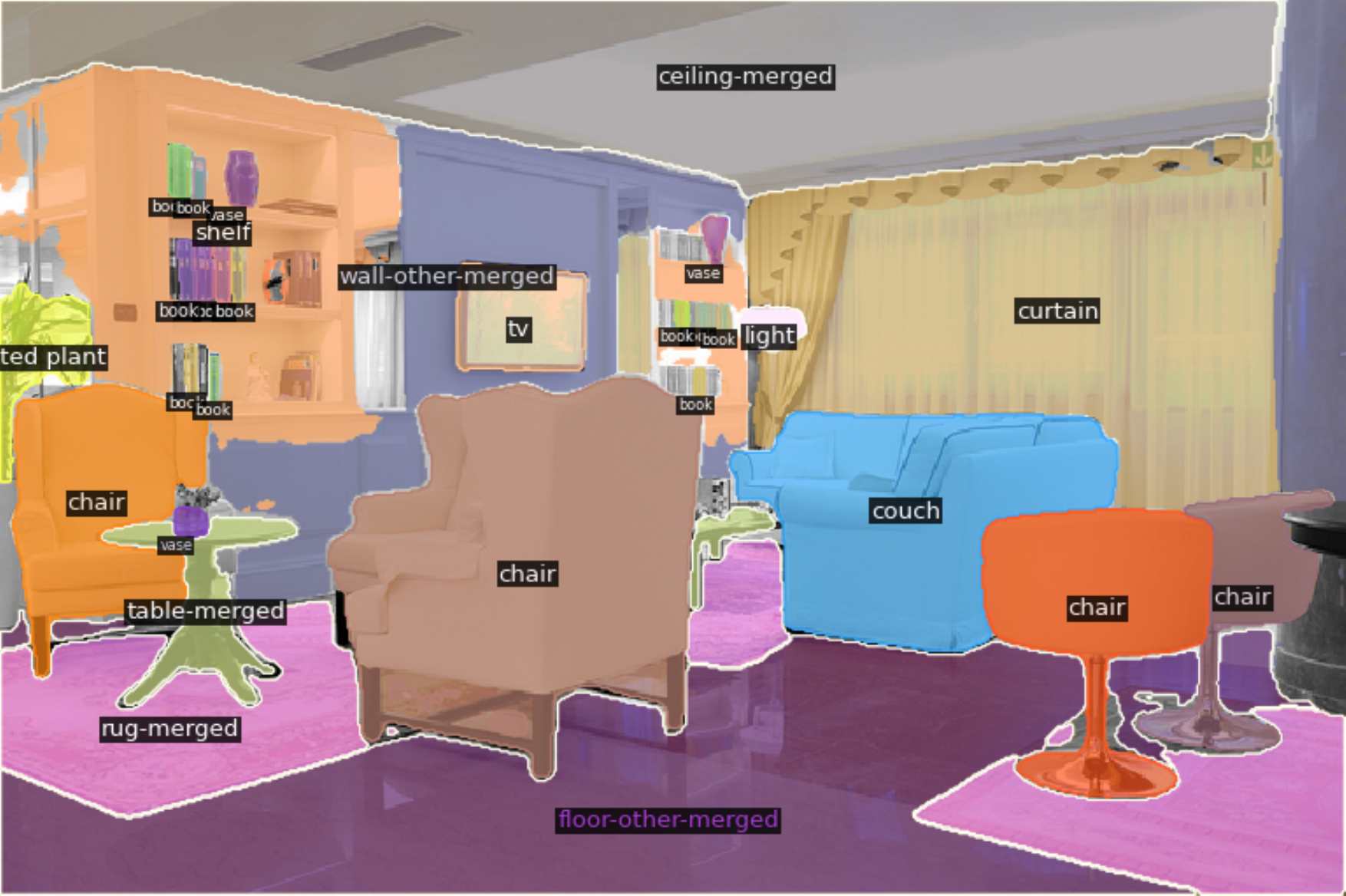} \\
    \raisebox{0.5ex}{\footnotesize (a) Input} & 
    \raisebox{0.5ex}{\footnotesize (b) Prediction (Instance Merge)} \\
    \includegraphics[width=0.225\textwidth, height=2.6cm]{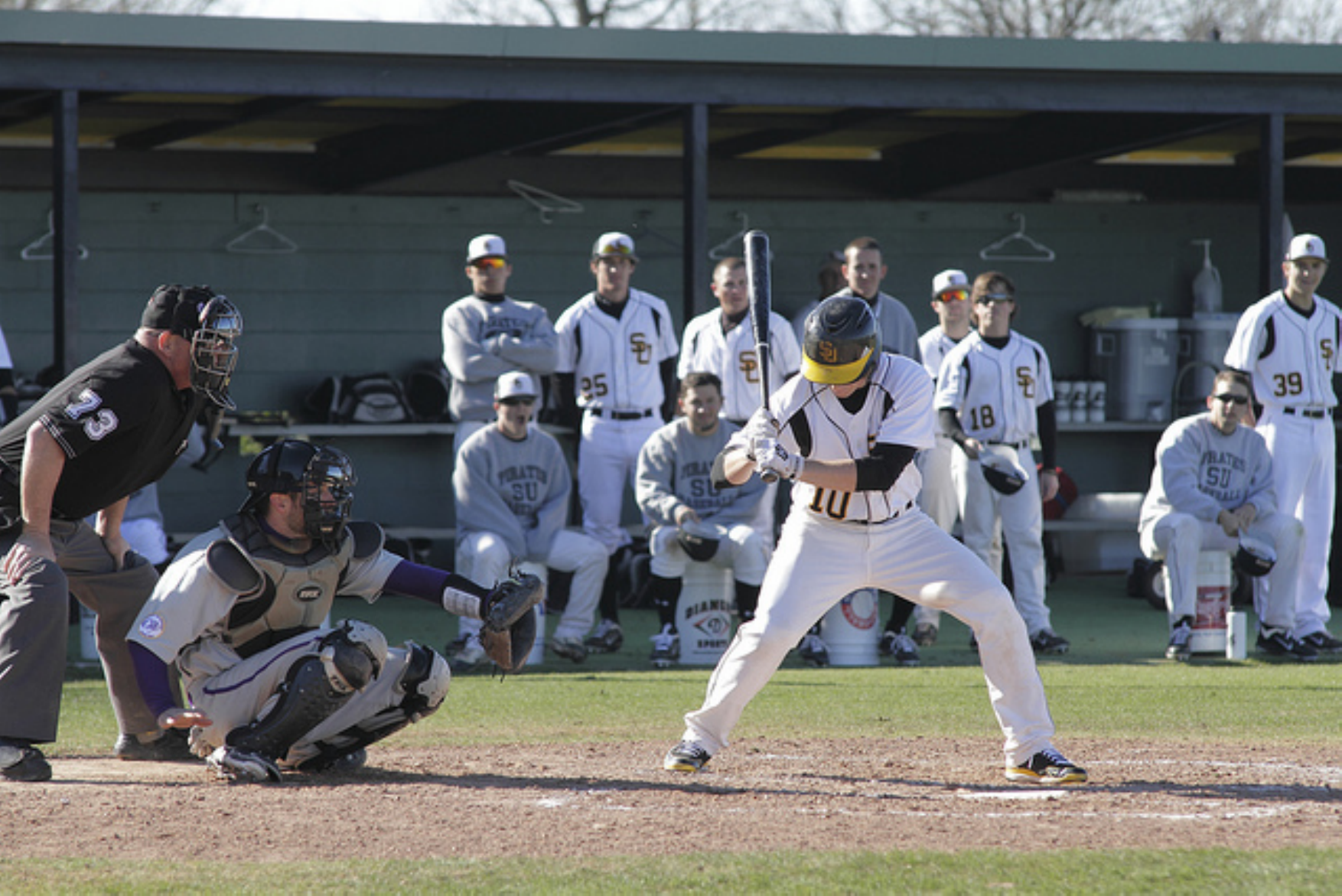} & 
    \includegraphics[width=0.225\textwidth, height=2.6cm]{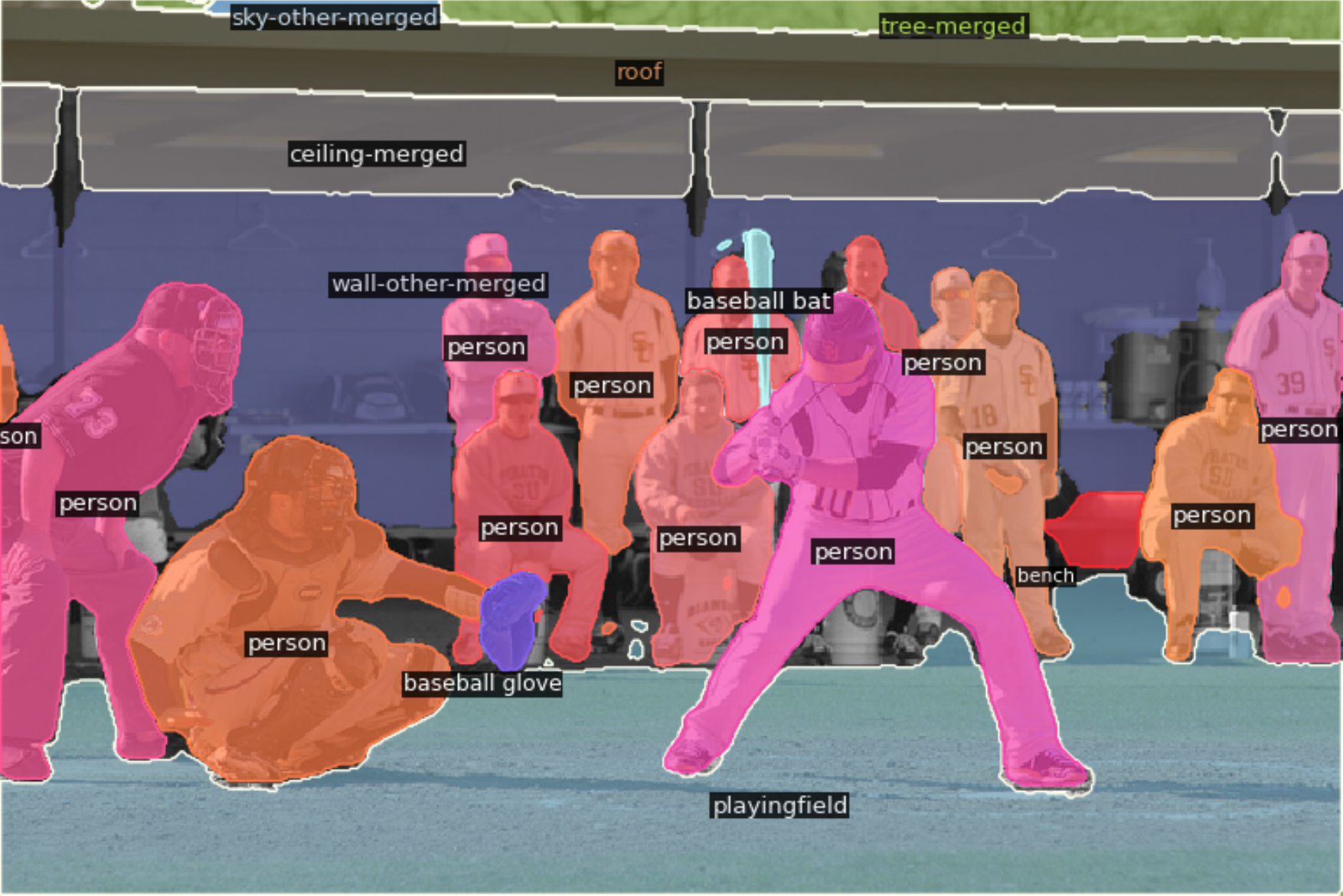} \\
    \raisebox{0.5ex}{\footnotesize (c) Input} & 
    \raisebox{0.5ex}{\footnotesize (d) Prediction (Instance Missing)}
    \end{tabular}
    \caption{Examples of occlusion-induced segmentation failures from COCO val2017~\cite{lin2014microsoft} using Mask2Former~\cite{cheng2022masked}. The left column shows the original input images, and the right column shows the segmentation predictions. In the top row, the brown chair in the center is actually two chairs, but the rear one is heavily occluded and is mistakenly merged with the front one in the prediction. In the bottom row, the person to the right of the blue baseball bat is missed due to severe occlusion, leaving only a dark region visible.}
   \label{fig:failures}
\end{figure}

Panoptic segmentation aims to achieve holistic scene understanding by jointly assigning both semantic and instance labels to every pixel~\cite{kirillov2019panoptic, kirillov2019fpn, cheng2020panoptic, li2021fully, cheng2021per, cheng2022masked, li2023mask, hu2023you, rashwan2024maskconver}. It shows a wide range of real-world applications, such as autonomous driving and robotics. In these scenarios, objects commonly appear in close proximity, leading to frequent occlusions. The human visual system can infer the identity and boundaries of occluded objects by leveraging prior experience and contextual cues~\cite{ullman2000high}. However, modern panoptic segmentation methods~\cite{cheng2021per, cheng2022masked, li2023mask} still suffer performance degradation in occluded scenes~\cite{wei2025cocoolac}. As illustrated in \autoref{fig:failures}, instances subject to occlusion are prone to boundary confusion or complete omission in the segmentation results from the representative model Mask2Former~\cite{cheng2022masked}.

This challenge arises mainly for two reasons: 1) Most vision benchmarks~\cite{deng2009imagenet, lin2014microsoft, cordts2016cityscapes} provide annotations only for the visible parts of each object, ignoring fine-grained occlusion details, such as depth order~\cite{qi2019amodal}, amodal mask~\cite{zhu2017semantic}, and occlusion ratio~\cite{milan2016mot16}, which results in limited supervision for learning occlusion-aware features. 2) State-of-the-art panoptic segmentation methods~\cite{cheng2021per, cheng2022masked, li2023mask} are typically transformer-based architectures that rely on the combination of an attention mechanism and position embedding to model global context with spatial priors. The attention mechanism infers pixel relationships by computing pairwise similarities, but in occluded scenes, incomplete context hinders its effectiveness. Similarly, widely used position embedding approaches~\cite{vaswani2017attention, shaw2018self, dosovitskiy2020image} normally encode content-agnostic spatial locations, preventing them from dynamically adapting to visual contexts.

In this paper, we propose PEMOLA, a novel occlusion-aware module designed for transformer-based panoptic segmentation. The key idea is to adapt position embedding to varying occlusions by integrating occlusion priors into the encoding process. To obtain these priors, we train an image-level occlusion classifier on the COCO-OLAC dataset~\cite{wei2025cocoolac}, which includes occlusion severity annotations (low, medium, and high) for each image. The classifier produces occlusion-level attention, serving as an occlusion prior that highlights the image regions most indicative of occlusion. All training images are preprocessed by blackening non-object regions, thereby guiding the classifier to focus more on foreground features and reducing interference from complex backgrounds. To substantiate the robustness of PEMOLA and promote occlusion-aware learning in the community, we further annotate the Cityscapes dataset~\cite{cordts2016cityscapes} following the same occlusion labelling rules as COCO-OLAC, which we term Cityscapes-OLAC, and validate the efficacy of PEMOLA on it.

PEMOLA leverages both occlusion priors in a complementary manner. The occlusion-level attention provides global spatial modulation, whereas the occlusion label is mapped to a learnable embedding vector to distribute the modulation across channels, ensuring that each channel captures occlusion in a distinct way. The combined occlusion-aware signals are applied to the position embedding via residual scaling, preserving the base structure while adaptively aligning with both the severity and distribution of occlusion. The overall approach is conceptually simple and lightweight, introducing only negligible computational overhead. Moreover, it can be seamlessly integrated into existing transformer-based panoptic segmentation methods without altering their architectures. 

Our contributions can be summarised as follows:

\begin{itemize}
\item We train an occlusion classifier on the COCO-OLAC dataset to generate occlusion-level attention and support occlusion prediction in other datasets.

\item We propose PEMOLA (Position Embedding Modulation with Occlusion-Level Attention), a lightweight module that incorporates occlusion priors into position embedding, achieving robust panoptic segmentation under varying occlusion conditions.

\item We manually annotate the Cityscapes dataset with image-level occlusion labels, termed Cityscapes-OLAC, using the same annotation protocol as COCO-OLAC. This provides a foundation for evaluating the cross-dataset generalization of PEMOLA and fosters future research on occlusion modelling.
\end{itemize}

\section{RELATED WORK}
\label{sec:relate}

\subsection{Panoptic Segmentation}
In early work, panoptic segmentation is formulated as a two-branch pipeline~\cite{kirillov2019panoptic, kirillov2019fpn}, with one branch predicting semantic categories and the other detecting instance identities. The outputs of both branches are combined using heuristic algorithms to produce the final result. Although effective, the non-learnable nature of heuristic algorithms can result in suboptimal segmentation performance. Following DETR~\cite{carion2020end}, recent panoptic segmentation methods~\cite{cheng2021per, cheng2022masked, li2023mask} have shifted towards transformer-based architectures, wherein the backbone extracts multi-scale features, the pixel decoder aggregates them, and the transformer decoder interacts with query embeddings to produce results. For example, MaskFormer~\cite{cheng2021per} formulates both semantic and instance segmentation as a mask classification problem. Mask2Former~\cite{cheng2022masked} proposes mask attention to constrain each query to attend only to the region within its prediction. Mask DINO~\cite{li2023mask} extends Mask2Former to object detection by incorporating denoising training~\cite{zhang2022dino}, achieving a unified framework for both detection and segmentation. However, these methods focus solely on modelling visible features, failing to capture explicit occlusion reasoning.

\subsection{Occlusion Modelling}
Occlusion is common but challenging. Numerous studies have explored occlusion modelling in different vision tasks. Bilayer Convolutional Network (BCNet)~\cite{ke2021deep} separates overlapping objects by modelling the image as an occluder layer and an occludee layer. Compositional Convolutional Networks (CompositionalNets)~\cite{kortylewski2021compositional} enhance convolutional neural networks with a compositional model that distinguishes object parts. Beyond modelling approaches, datasets have also been developed to benchmark occlusion handling. OVIS~\cite{qi2022occluded}, MOSE~\cite{ding2023mose}, and MOSEv2~\cite{ding2025mosev2} provide large-scale video datasets with densely annotated instances for video instance segmentation. COCOA~\cite{zhu2017semantic} extends COCO~\cite{lin2014microsoft} with amodal annotations covering both visible and occluded regions for amodal segmentation. These annotations are all instance-level, focusing on object masks and boundaries. In contrast, COCO-OLAC~\cite{wei2025cocoolac} introduces image-level occlusion labels, avoiding the need for fine-grained annotations. This enables us to train an occlusion classifier that learns discriminative features and generalises the occlusion labelling scheme to other datasets.

\begin{figure*}[t]
  \centering
   \includegraphics[width=0.9\linewidth]{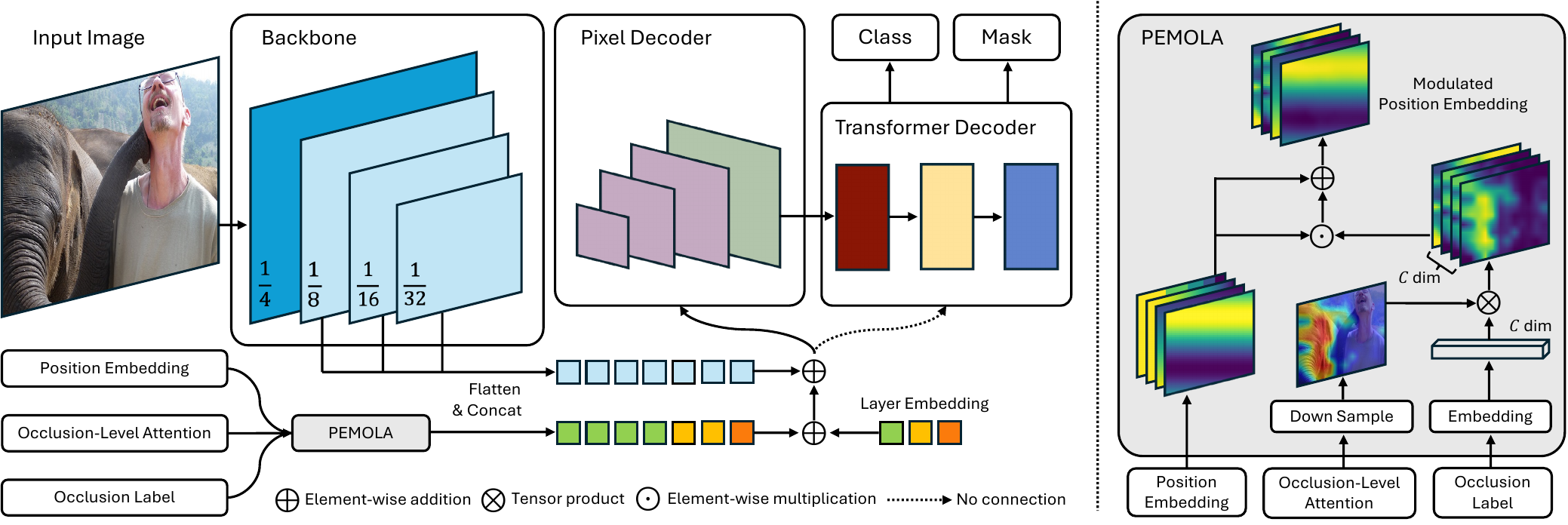}
   \caption{Overview of the proposed PEMOLA module. PEMOLA modulates position embedding using two occlusion priors: the occlusion-level attention extracted from the occlusion classifier highlights spatial regions affected by occlusion, while the learnable occlusion label embeddings encode global occlusion-level semantics to guide modulation. These two priors jointly adjust the positional representation in a residual manner, allowing the model to preserve spatial consistency while adaptively enhancing representations in occluded regions.}
   \label{fig:pemola}
\end{figure*}

\section{METHOD}
We introduce PEMOLA, a position embedding modulation module that enhances panoptic segmentation under occlusion. To better understand the motivation, we first revisit the concept of position embedding. Next, we present the implementation of the occlusion classifier. We then describe how occlusion priors are utilised by PEMOLA to modulate position embedding. Finally, we introduce the Cityscapes-OLAC dataset, which supports the cross-dataset evaluation of PEMOLA.

\subsection{Preliminaries: Position Embedding}
A self-attention mechanism computes pairwise dependencies in a symmetric manner, which makes it inherently permutation-invariant. To this end, the vanilla transformer~\cite{vaswani2017attention} introduced the 1D sinusoidal positional encoding to encode positional relations as:

\begin{equation}
\begin{aligned}
PE_{(p,2i)} &= \sin\!\left(p/{10000^{\frac{2i}{d}}}\right), \\
PE_{(p,2i+1)} &= \cos\!\left(p/{10000^{\frac{2i}{d}}}\right),
\end{aligned}
\label{eq:pe_1d}
\end{equation}

\noindent 
where $p$ is a token index, $d$ is the embedding dimension, and $i$ is the dimension index.

DETR~\cite{carion2020end} extends the 1D position embedding to a 2D formulation that aligns with the spatial grid of the image:

\begin{align}
PE(x, y) = PE(x) \oplus PE(y),
\label{eq:pe_2d}
\end{align}

\noindent
where $(x, y)$ are the spatial coordinates, and $\oplus$ is the concatenation operation. However, such positional encoding is fixed, limiting its flexibility under diverse visual conditions.

Subsequent works have developed various position embedding variants, including learnable~\cite{carion2020end}, relative~\cite{liu2021swin}, and rotary forms~\cite{su2024roformer}. Nevertheless, these approaches mainly refine the structural formulation of position embedding without incorporating image features, such as occlusion.

\subsection{Occlusion Classifier}
Most segmentation datasets do not provide occlusion annotations, leaving the model without direct supervision to understand spatial relationships among objects. To enable occlusion analysis, the COCO-OLAC dataset~\cite{wei2025cocoolac} was introduced as an extension of the COCO dataset~\cite{lin2014microsoft}, in which each image is manually labelled with a low, medium, or high level of occlusion. Although these annotations are relatively coarse, we argue that training an occlusion classifier is sufficient to learn discriminative features (See the attention map in~\Cref{fig:pe_visual}.).

Given an input image $\mathbf{I} \in \mathbb{R}^{H \times W \times 3}$ with a corresponding occlusion label $\mathbf{L} \in \{0,1,2\}$ (low, medium, and high), the classifier $\mathcal{Z}$ predicts $\mathbf{P} \in [0,1]^3$, where $\sum_i \mathbf{P}_i = 1$, indicating the confidence for each occlusion category. As the occlusion labels are annotated solely based on the instances, all training images are preprocessed by blackening non-object regions. The training is optimised using a standard binary-cross-entropy loss $\mathcal{L}_{\text{cls}}$. We conduct comprehensive evaluations using both ResNet~\cite{he2016deep} and Swin Transformer~\cite{liu2021swin} with different variants. On the black-background validation set, our classifier consistently achieves strong performance, demonstrating that even a simple classification network can effectively learn complex occlusion representations. Note that we finally adopt the Swin-L model~\cite{liu2021swin} as our occlusion classifier.

\subsection{PEMOLA}
Current transformer-based panoptic segmentation methods rely on position embedding to encode positional relationships. However, position embedding is content-independent, which limits its adaptability to complex scene conditions. To alleviate this issue, we propose PEMOLA, a lightweight module that modulates position embedding through the incorporation of both occlusion-level attention and the occlusion label embedding, thereby enhancing spatial awareness under occlusion, as shown in \Cref{fig:pemola}. 

\noindent
\textbf{Occlusion-Level Attention} aims to capture the spatial distribution of occlusion within an image. It is extracted from the occlusion classifier $\mathcal{Z}$ using Grad-CAM (Gradient-weighted Class Activation Mapping)~\cite{selvaraju2017grad}, which computes the gradients of the prediction $\mathbf{P}_t$ with respect to the feature layer $\mathbf{F}^k$ (we select the last normalization output in the final block of $\mathcal{Z}$) to highlight the regions that are most relevant to the occlusion-level decision. This process can be formulated in two steps. 

First, Grad-CAM averages the gradients of the target prediction $\mathbf{P}_t$ over all spatial locations.

\begin{equation}
\mathbf{W}_k^t = \frac{1}{hw} 
\sum_{i=1}^{h} \sum_{j=1}^{w} 
\frac{\partial \mathbf{P}_t}{\partial \mathbf{F}_{ij}^k},
\end{equation}

\noindent
where $\mathbf{W}_k^t$ denotes the importance weight of feature layer $\mathbf{F}^k$ for class $t$. $h$ and $w$ denote the height and width of the feature map. $i$ and $j$ denote the spatial location.

Next, the occlusion-level attention $\mathbf{O}_a \in \mathbb{R}^{h \times w}$ is produced by combining all weighted feature maps, followed by ReLU to retain only the positive regions.

\begin{equation}
\mathbf{O}_a = \mathrm{ReLU}\!\left(\sum_k \mathbf{W}_k^t \mathbf{F}^k \right).
\end{equation}

\noindent
\textbf{Occlusion Label Embedding} serves as a learnable representation of occlusion levels to modulate the channel-wise weights of the position embedding. It is generated by an embedding function $\mathcal{E}:\{0,1,2\}\rightarrow\mathbb{R}^{\mathcal{C}}$, which maps each occlusion label $\mathbf{L} \in \{0,1,2\}$ (low, medium, and high) to a $\mathcal{C}$-dimensional vector through $\mathcal{E}(\mathbf{L})$.

\noindent
\textbf{Occlusion-Aware Position Embedding} is obtained by combining the two occlusion priors into a unified modulation signal, as shown on the right in~\Cref{fig:pemola}. The occlusion-level attention $\mathbf{O}_a$ is first interpolated to match the spatial resolution $(h, w)$ of the position embedding and then expanded to the shape $(b, 1, h, w)$. The occlusion label embedding $\mathbf{O}_l$ is expanded to $(b, \mathcal{C}, 1, 1)$ to align with the channel dimension. Finally, the original position embedding is modulated via

\begin{equation}
\mathbf{E}_{pos}^m = \mathbf{E}_{pos} \odot (1 + \mathbf{O}_a \otimes \mathbf{O}_l),
\end{equation}

\noindent
where $\mathbf{E}_{pos}^m \in \mathbb{R}^{b\times \mathcal{C}\times h\times w}$ denotes the modulated position embedding, $\mathbf{E}_{pos} \in \mathbb{R}^{b\times \mathcal{C}\times h\times w}$ denotes the original position embedding, $\otimes$ denotes broadcast multiplication, and $\odot$ denotes element-wise multiplication. This formulation preserves the structural integrity of the position embedding and enables adaptive modulation under different occlusion conditions.

\begin{table}[!t]
\caption{Occlusion Classifier on the split COCO-OLAC dataset~\cite{wei2025cocoolac}. Black denote background blackening. acc@1 indicates the top-1 classification accuracy.}
\centering
\begin{adjustbox}{width={0.43\textwidth}, keepaspectratio}
\begin{tabular}{@{\hspace{4pt}}l|c|c c|c c@{\hspace{4pt}}}
\toprule
Model & \makecell{Image\\Size} & \makecell{val\\acc@1} & \makecell{test\\acc@1} & \makecell{val$^\text{black}$\\acc@1} & \makecell{test$^\text{black}$\\acc@1} \\ 
\midrule
R-50~\cite{he2016deep} & $\text{224}^2$ & 53.6 & 51.9 & 72.2 & 70.3 \\
R-101 & $\text{224}^2$ & 55.4 & 53.2 & 72.7 & 70.8 \\
\midrule
Swin-T~\cite{liu2021swin} & $\text{224}^2$ & 61.7 & 60.8 & 72.1 & 71.6 \\
Swin-S & $\text{224}^2$ & 63.5 & 62.3 & 73.5 & 71.4 \\
Swin-B & $\text{224}^2$ & 61.8 & 59.9 & 73.9 & 71.7 \\
Swin-L &$\text{224}^2$ & 62.8 & 62.1 & 74.3 & 73.0 \\
\midrule
Swin-B & $\text{384}^2$ & 63.1 & 62.6 & 77.7 & 75.3 \\
Swin-L & $\text{384}^2$ & \textbf{65.2} & \textbf{63.6} & \textbf{78.2} & 75.3 \\
\bottomrule
\end{tabular}
\end{adjustbox}
\label{table:coco_class}
\end{table}

\begin{table}[!t]
\caption{Occlusion Classifer on the newly annotated Cityscapes dataset~\cite{cordts2016cityscapes}. Black denote background blackening. acc@1 indicates the top-1 classification accuracy.}
\centering
\begin{adjustbox}{width={0.43\textwidth}, keepaspectratio}
\begin{tabular}{@{\hspace{4pt}}l|c|c c|c c@{\hspace{4pt}}}
\toprule
Model & \makecell{Image\\Size} & \makecell{train\\acc@1} & \makecell{val\\acc@1} & \makecell{train$^\text{black}$\\acc@1} & \makecell{val$^\text{black}$\\acc@1} \\ 
\midrule
Swin-L~\cite{liu2021swin} & $\text{384}^2$ & 68.5 & 71.5 & 84.8 & 86.2 \\
\bottomrule
\end{tabular}
\end{adjustbox}
\label{table:city_class}
\end{table}

We incorporate the modulated position embedding only into the pixel decoder rather than the transformer decoder. This is because the transformer decoder operates on deeper and more abstract features, whose representation space is less aligned with that of the occlusion classifier. The qualitative analysis presented in~\Cref{table:pemola_location} further confirms this design choice.

\subsection{Cityscapes-OLAC Dataset}
To evaluate the cross-dataset generalisation ability of PEMOLA, we manually annotated the Cityscapes dataset, following the same occlusion labelling protocol as COCO-OLAC~\cite{wei2025cocoolac}, which we refer to as Cityscapes-OLAC. This dataset contains 2,975 training images, with 2,521 labelled as high occlusion, 343 as medium, and 111 as low. The validation set contains 500 images, with 428 labelled as high occlusion, 53 as medium, and 19 as low. Compared with COCO-OLAC, Cityscapes-OLAC exhibits a highly imbalanced distribution of occlusion levels, along with distinct scene layouts and occlusion patterns. For instance, because Cityscapes-OLAC consists exclusively of urban street scene images, occlusions primarily result from vehicles, pedestrians, and various roadside elements, such as poles and traffic lights.

\section{EXPERIMENTS}
We conduct extensive experiments following the same settings as previous works~\cite{he2016deep, liu2021swin, cheng2022masked, li2023mask} for a fair comparison. The learning rate is adjusted for the occlusion classifier to achieve optimal performance. All models adopt ResNet-50~\cite{he2016deep} as the backbone, excluding the occlusion classifier.

\begin{table}[!t]
\caption{Quantitative comparisons across low, mid, and high occlusion levels on the COCO-OLAC dataset~\cite{wei2025cocoolac}.}
\centering
\begin{adjustbox}{width={0.48\textwidth}, keepaspectratio}
\begin{tabular}{@{\hspace{4pt}}l|c|c c c|c c@{\hspace{4pt}}}
\toprule
Model & Level & PQ & $\text{PQ}^{\text{Th}}$ & $\text{PQ}^{\text{St}}$ & $\text{AP}_{\text{pan}}^{\text{Th}}$ & $\text{mIoU}_{\text{pan}}$ \\ 
\midrule
\multicolumn{7}{l}{\textit{Mask2Former}~\cite{cheng2022masked}} \\
\midrule
Base & \multirow{2}{*}{low} & 46.9 & 53.1 & 37.8 & 45.5 & 52.1 \\
+PEMOLA (ours) &  & \textbf{47.5} & 53.1 & \textbf{39.2} & \textbf{47.9} & \textbf{55.1} \\
\midrule
Base & \multirow{2}{*}{mid} & 42.1 & 46.9 & 35.0 & 32.7 & \textbf{54.5} \\
+PEMOLA (ours) &  & \textbf{42.9} & \textbf{47.6} & \textbf{35.9} & \textbf{32.9} & 54.3 \\
\midrule
Base & \multirow{2}{*}{high} & 35.2 & 37.9 & 31.5 & 24.0 & \textbf{50.4} \\
+PEMOLA (ours) &  & \textbf{35.7} & \textbf{38.1} & \textbf{32.2} & 24.0 & 49.6 \\
\midrule
\multicolumn{1}{l}{\textit{Mask DINO}~\cite{li2023mask}} \\
\midrule
Base & \multirow{2}{*}{low} & 48.6 & 54.6 & 39.6 & 48.2 & 51.3 \\
+PEMOLA (ours) &  & \textbf{51.9} & \textbf{58.8} & \textbf{41.6} & \textbf{50.0} & \textbf{55.3} \\
\midrule
Base & \multirow{2}{*}{mid} & 45.5 & 50.7 & 37.6 & 36.1 & 53.4 \\
+PEMOLA (ours) &  & 45.5 & 50.7 & \textbf{37.7} & \textbf{36.2} & \textbf{54.2} \\
\midrule
Base & \multirow{2}{*}{high} & 38.4 & 41.4 & 33.8 & 27.4 & 49.7 \\
+PEMOLA (ours) &  & \textbf{39.4} & \textbf{42.9} & \textbf{34.2} & \textbf{28.3} & \textbf{51.6} \\
\bottomrule
\end{tabular}
\end{adjustbox}
\label{table:panoptic_level}
\end{table}

\begin{table}[!t]
\caption{Panoptic Segmentation comparison on the COCO-OLAC dataset~\cite{wei2025cocoolac}. ${\dagger}$ denotes the retrained model.}
\centering
\begin{adjustbox}{width={0.46\textwidth}, keepaspectratio}
\begin{tabular}{@{\hspace{4pt}}l|c c c|c c@{\hspace{4pt}}}
\toprule
Model & $\text{PQ}$ & $\text{PQ}^{\text{Th}}$ & $\text{PQ}^{\text{St}}$ & $\text{AP}_{\text{pan}}^{\text{Th}}$ & $\text{mIoU}_{\text{pan}}$ \\ 
\midrule
Panoptic FPN~\cite{kirillov2019fpn} & 33.3 & 39.1 & 24.6 & 26.9 & 36.6 \\
Panoptic FCN~\cite{li2021fully} & 33.0 & 37.7 & 26.0 & 23.1 & 34.8 \\
Panoptic-DeepLab~\cite{cheng2020panoptic} & 27.5 & 28.9 & 25.3 & 13.8 & - \\
YOSO~\cite{hu2023you} & 37.1 & 40.6 & 31.9 & - & - \\
\midrule
Mask2Former$^{\dagger}$~\cite{cheng2022masked} & 40.7 & 44.5 & 35.0 & 30.0 & 54.2 \\
+ PEMOLA (ours) & \textbf{41.5} & \textbf{45.2} & \textbf{35.9} & \textbf{30.4} & \textbf{54.8} \\
\midrule
Mask DINO$^{\dagger}$~\cite{li2023mask} & 44.0 & 48.5 & 37.3 & 33.5 & 53.4 \\
+ PEMOLA (ours) & \textbf{44.8} & \textbf{49.4} & \textbf{37.8} & \textbf{34.2} & \textbf{55.3} \\
\bottomrule
\end{tabular}
\end{adjustbox}
\label{table:panoptic_olac}
\end{table}

\begin{table}[!t]
\caption{Panoptic Segmentation comparison on the newly annotated Cityscapes-OLAC dataset. ${\dagger}$ denotes the retrained model.}
\centering
\begin{adjustbox}{width={0.46\textwidth}, keepaspectratio}
\begin{tabular}{@{\hspace{4pt}}l|c c c|c c@{\hspace{4pt}}}
\toprule
Model & $\text{PQ}$ & $\text{PQ}^{\text{Th}}$ & $\text{PQ}^{\text{St}}$ & $\text{AP}_{\text{pan}}^{\text{Th}}$ & $\text{mIoU}_{\text{pan}}$ \\ 
\midrule
Panoptic FPN~\cite{kirillov2019fpn} & 58.1 & 52.0 & 62.5 & 33.0 & 75.7 \\
Panoptic FCN~\cite{li2021fully} & 59.6 & 52.1 & 65.1 & - & - \\
Panoptic-DeepLab~\cite{cheng2020panoptic} & 60.3 & 51.1 & 67.0 & 32.2 & 78.2 \\
YOSO~\cite{hu2023you} & 59.7 & 51.0 & 66.1 & - & - \\
\midrule
Mask2Former$^{\dagger}$~\cite{cheng2022masked} & 61.5 & 54.0 & 66.9 & 35.2 & 76.1 \\
+ PEMOLA (ours) & \textbf{62.3} & \textbf{55.4} & \textbf{67.2} & \textbf{38.5} & \textbf{77.4} \\
\bottomrule
\end{tabular}
\end{adjustbox}
\label{table:panoptic_city}
\end{table}

\subsection{Datasets and Evaluation Metrics}
We evaluate PEMOLA on Panoptic Segmentation using two challenging datasets: COCO-OLAC~\cite{wei2025cocoolac} and the newly annotated Cityscapes-OLAC. For training the occlusion classifier, we randomly split the COCO-OLAC training set into 90\% for training and 10\% for validation. The original COCO-OLAC validation set is used as the test set.

For panoptic segmentation, we adopt standard evaluation metrics, including PQ, $\text{PQ}^{\text{Th}}$, and $\text{PQ}^{\text{St}}$, which measure the overall, ``thing'', and ``stuff'' segmentation quality, respectively. In addition, we report $\text{AP}_{\text{pan}}^{\text{Th}}$ for instance segmentation and $\text{mIoU}_{\text{pan}}$ for semantic segmentation, both computed using panoptic segmentation annotations.

\subsection{Main Results}

\noindent
\textbf{Occlusion Classifier.} We compare various backbones for occlusion classifiers on the split COCO-OLAC dataset. As shown in \Cref{table:coco_class}, the Swin Transformer~\cite{liu2021swin} consistently outperforms CNN-based methods (ResNet-50 and ResNet-101). Increasing the model size and input resolution further improves accuracy. Moreover, models evaluated on black-background images yield higher accuracies, with the Swin-L achieving the best performance of 78.2\% on the validation set. This indicates that the background context can interfere with occlusion reasoning. We also evaluate the occlusion classifier on the Cityscapes-OLAC dataset in \Cref{table:city_class}. The Swin-L reaches 86.2\% accuracy on the black-background validation set, demonstrating strong cross-dataset generalisation.

\begin{figure}[t]
    \centering
    \setlength{\tabcolsep}{0.2pt}
    \renewcommand{\arraystretch}{0.1}
    \begin{tabular}{@{}cccccccc@{}}
    \includegraphics[width=1.08cm, height=1.08cm]{fig/pe/low/244019.pdf} & 
    \includegraphics[width=1.08cm, height=1.08cm]{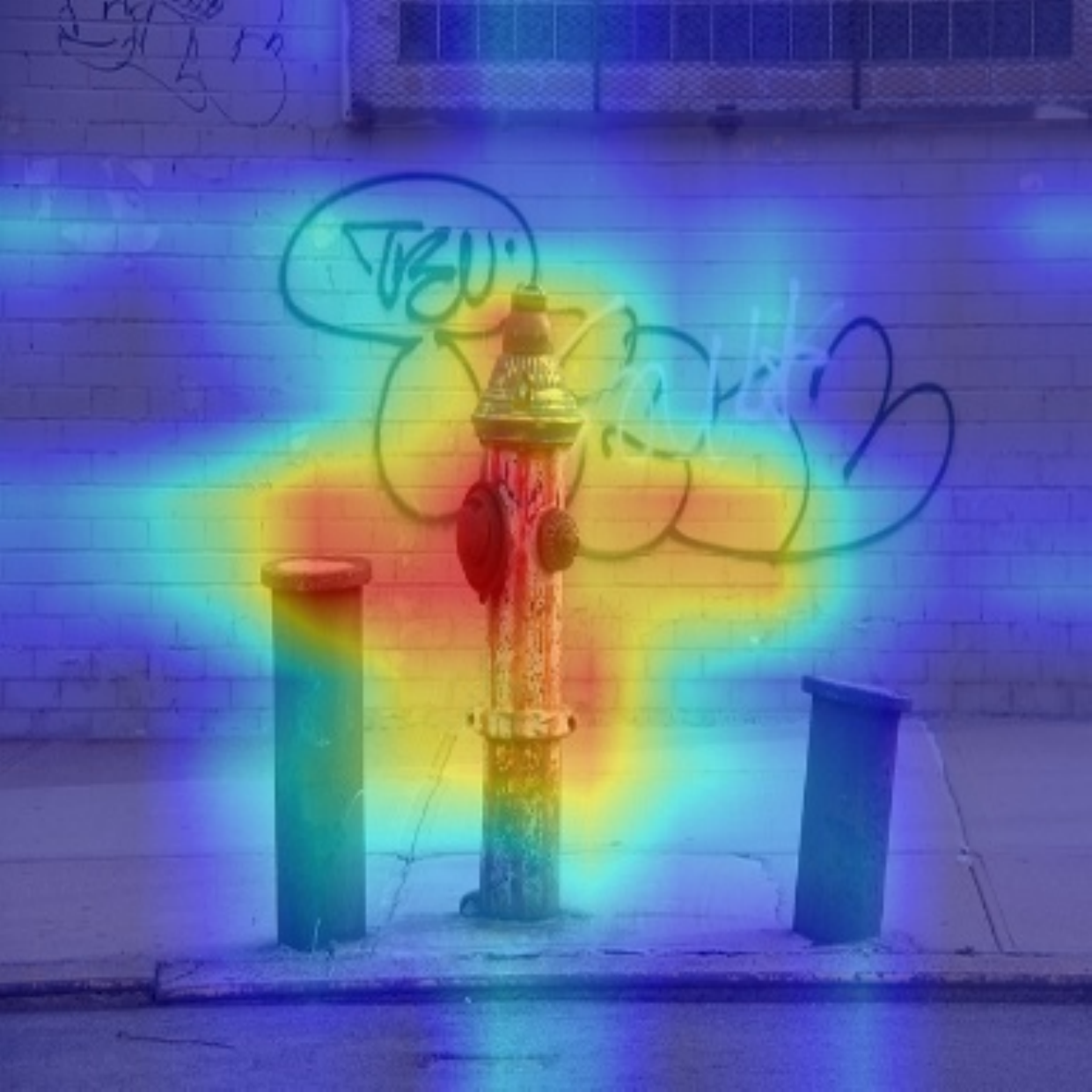} & 
    \includegraphics[width=1.08cm, height=1.08cm]{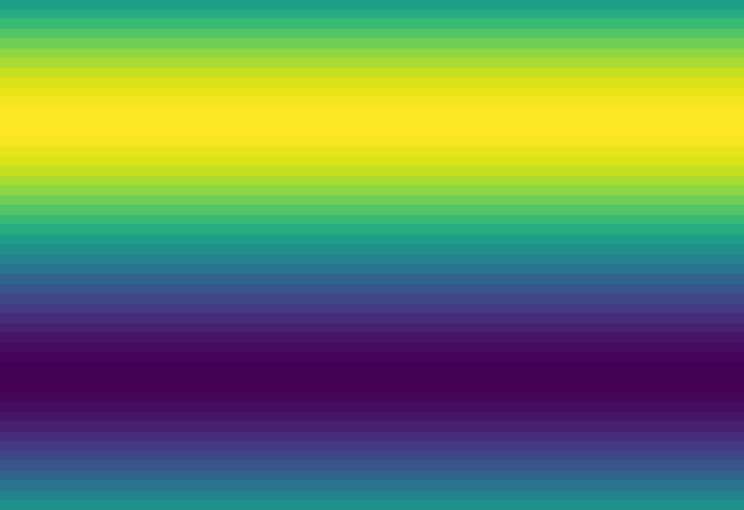} & 
    \includegraphics[width=1.08cm, height=1.08cm]{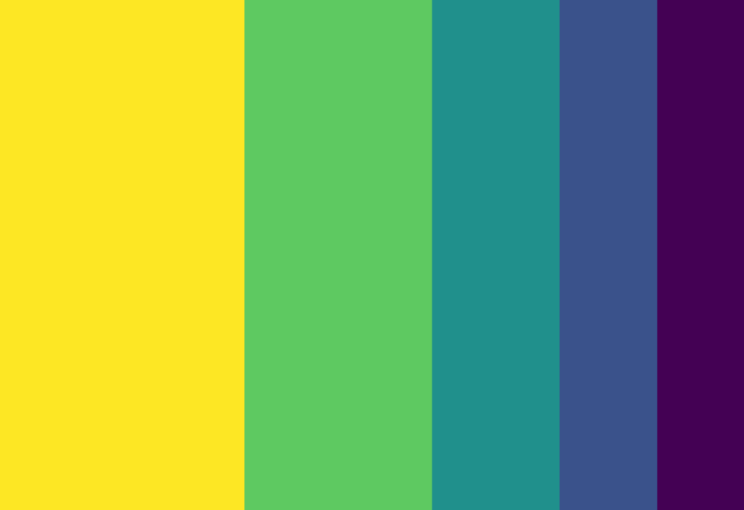} & 
    \includegraphics[width=1.08cm, height=1.08cm]{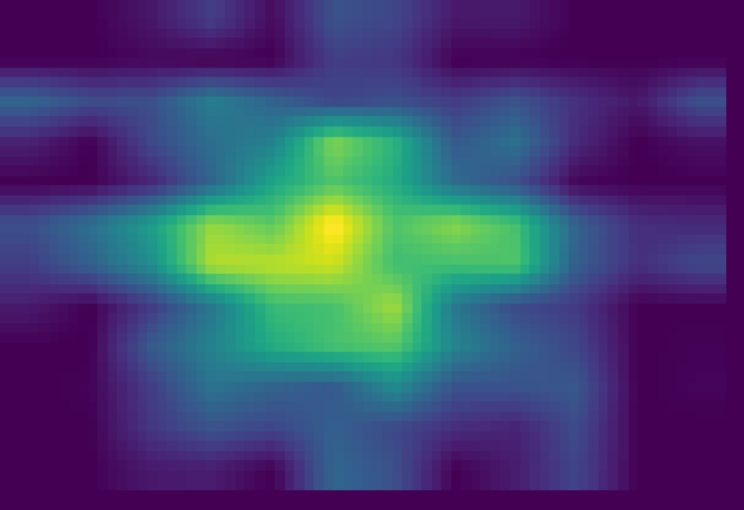} & 
    \includegraphics[width=1.08cm, height=1.08cm]{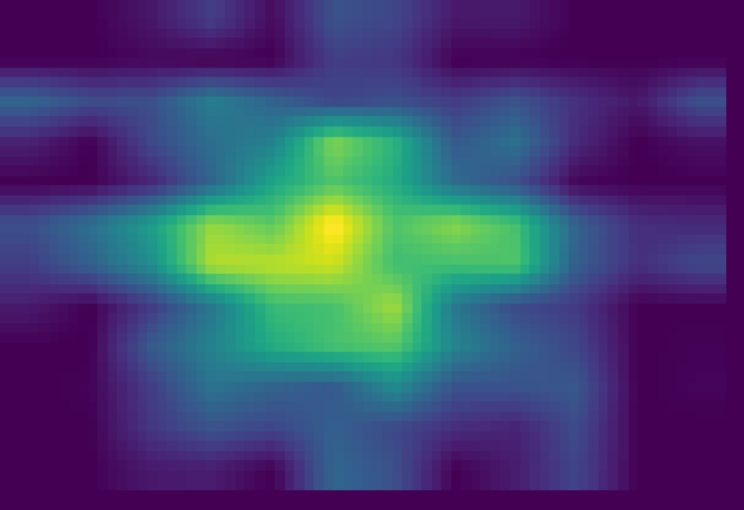} & 
    \includegraphics[width=1.08cm, height=1.08cm]{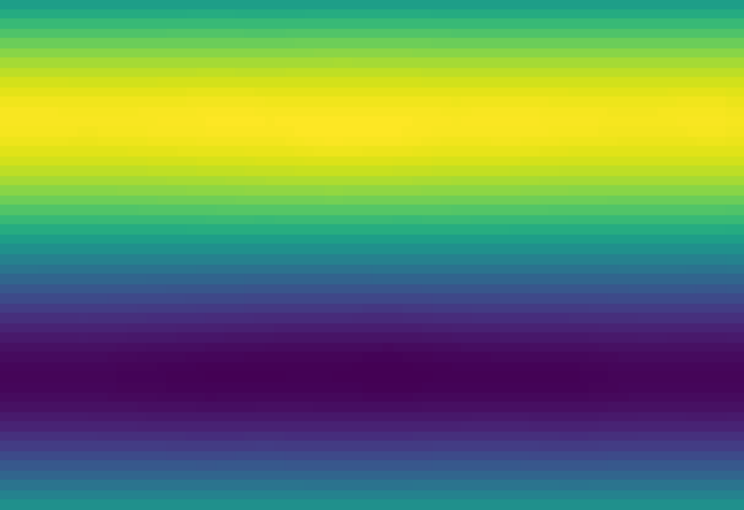} & 
    \includegraphics[width=1.08cm, height=1.08cm]{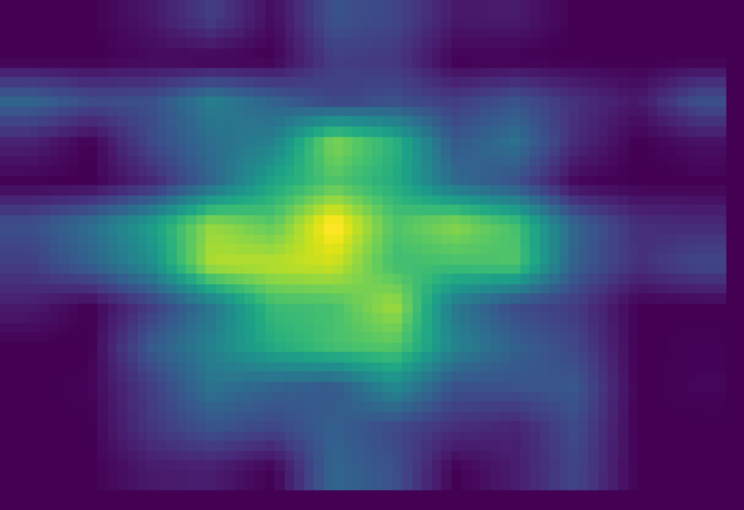} \\ 
    \includegraphics[width=1.08cm, height=1.08cm]{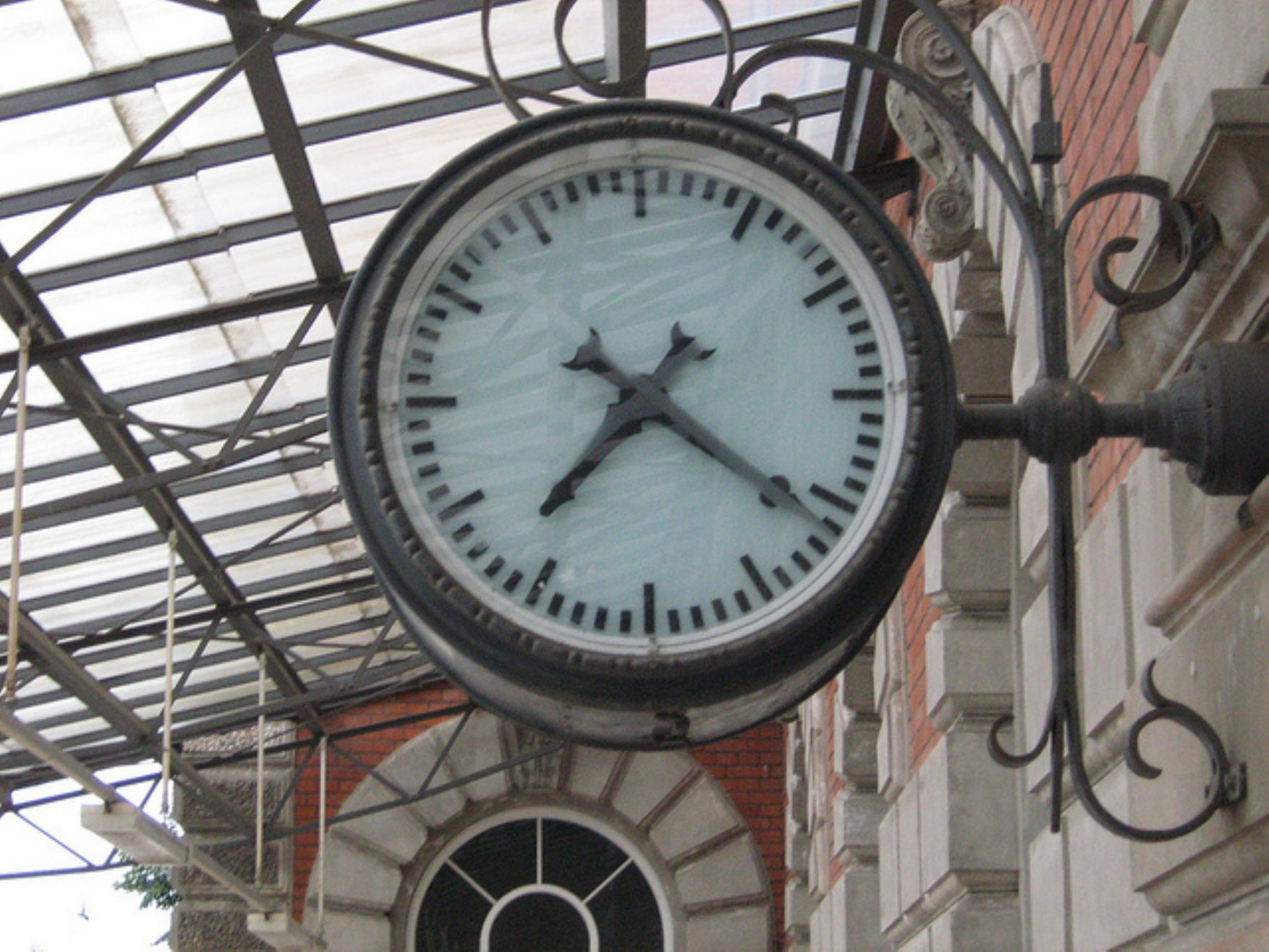} & 
    \includegraphics[width=1.08cm, height=1.08cm]{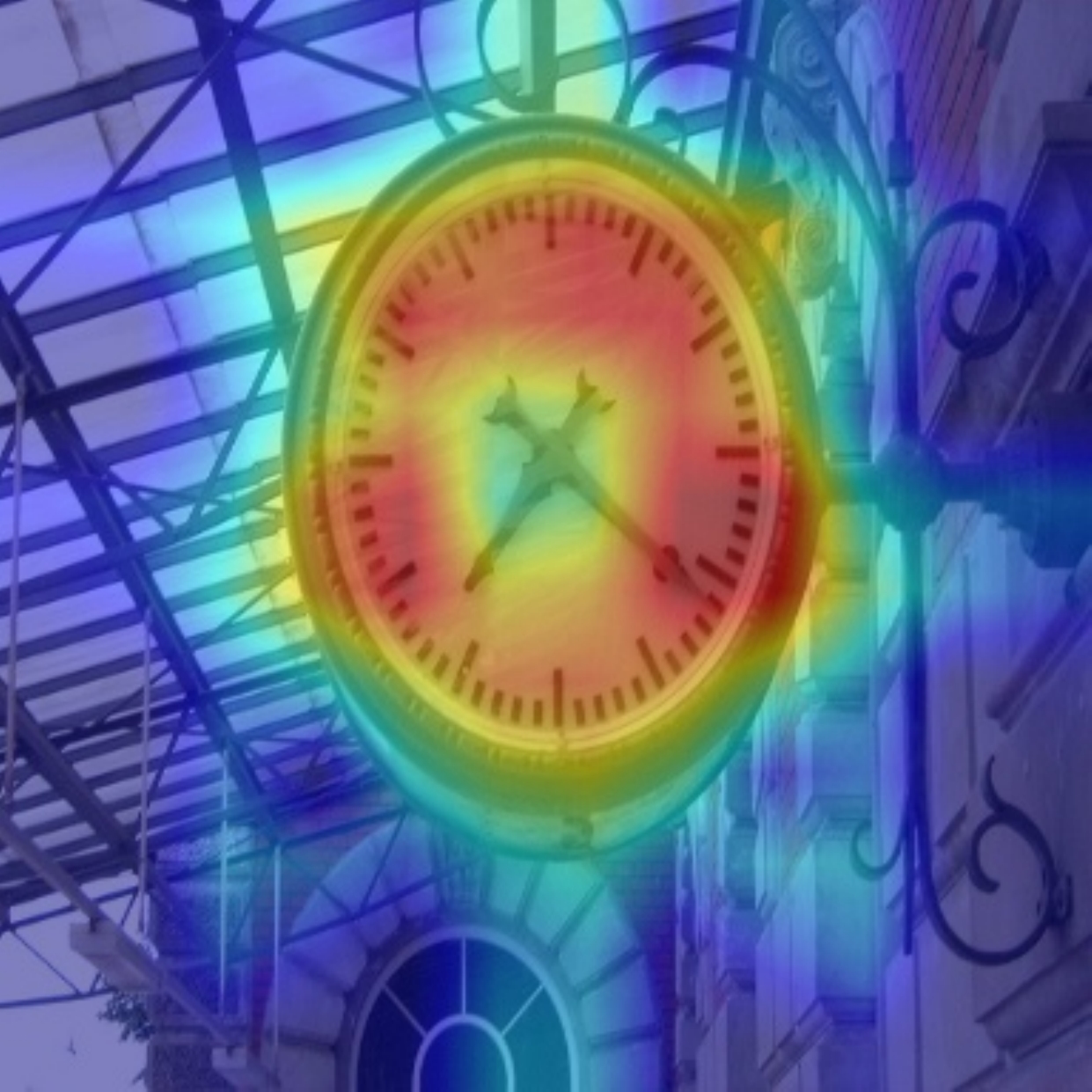} & 
    \includegraphics[width=1.08cm, height=1.08cm]{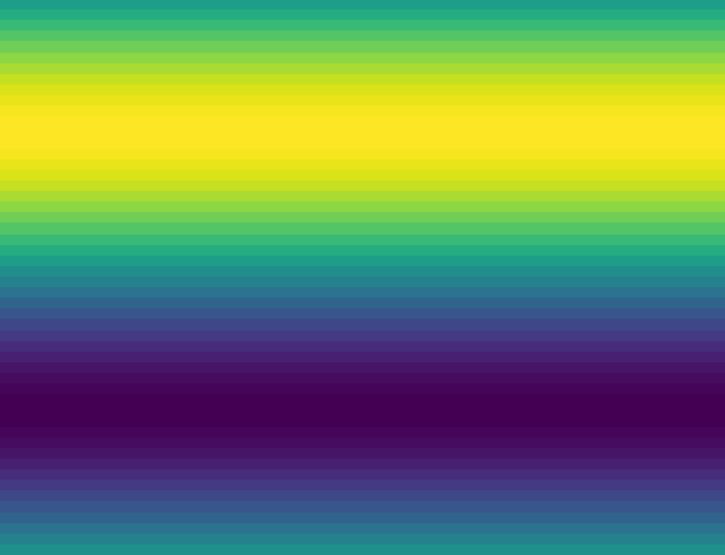} & 
    \includegraphics[width=1.08cm, height=1.08cm]{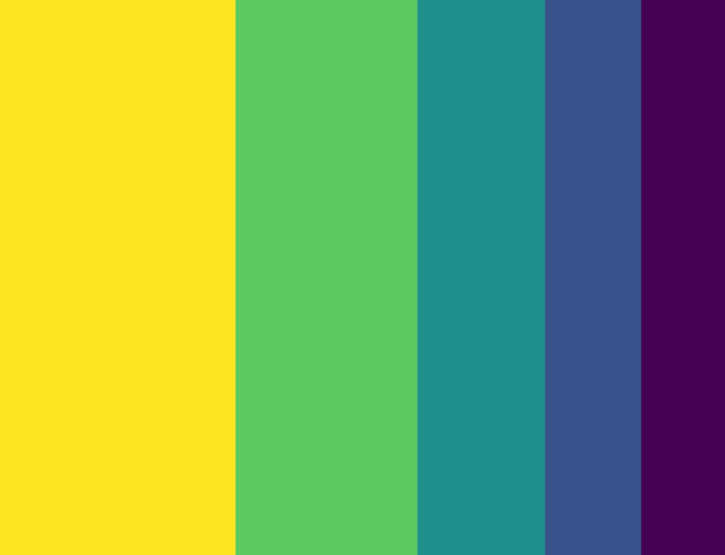} & 
    \includegraphics[width=1.08cm, height=1.08cm]{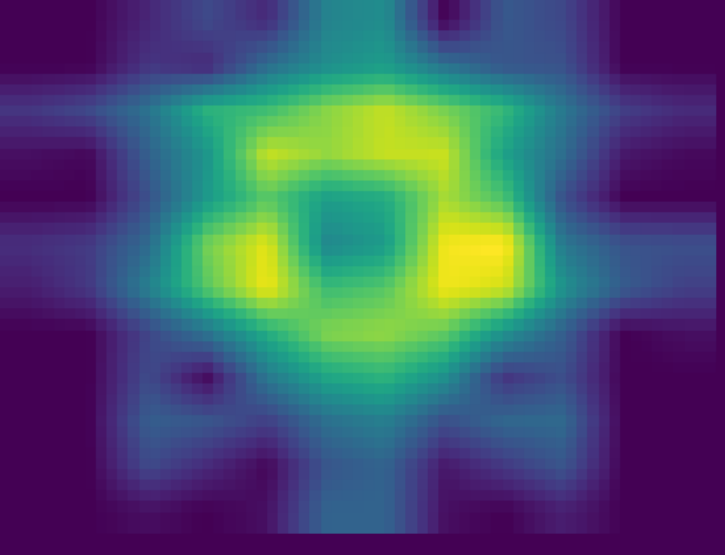} & 
    \includegraphics[width=1.08cm, height=1.08cm]{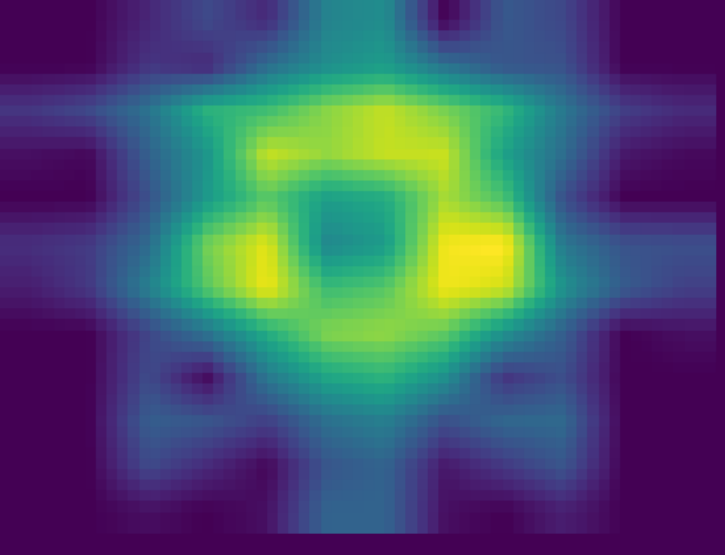} & 
    \includegraphics[width=1.08cm, height=1.08cm]{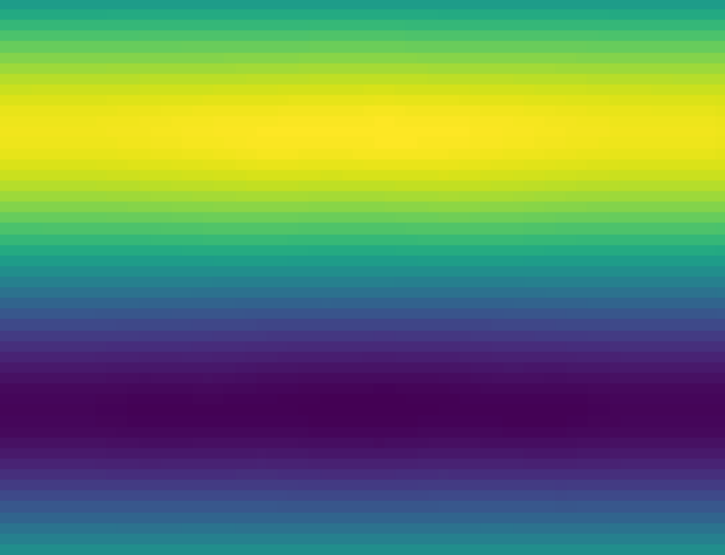} & 
    \includegraphics[width=1.08cm, height=1.08cm]{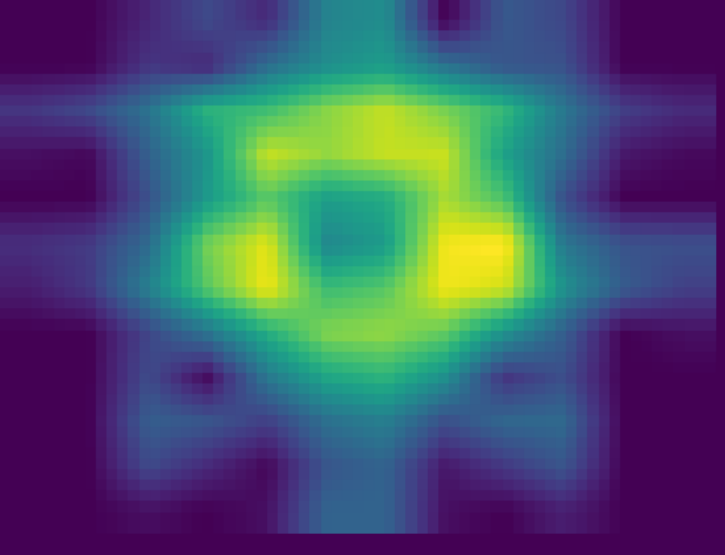} \\ 
    \includegraphics[width=1.08cm, height=1.08cm]{fig/pe/mid/31269.pdf} & 
    \includegraphics[width=1.08cm, height=1.08cm]{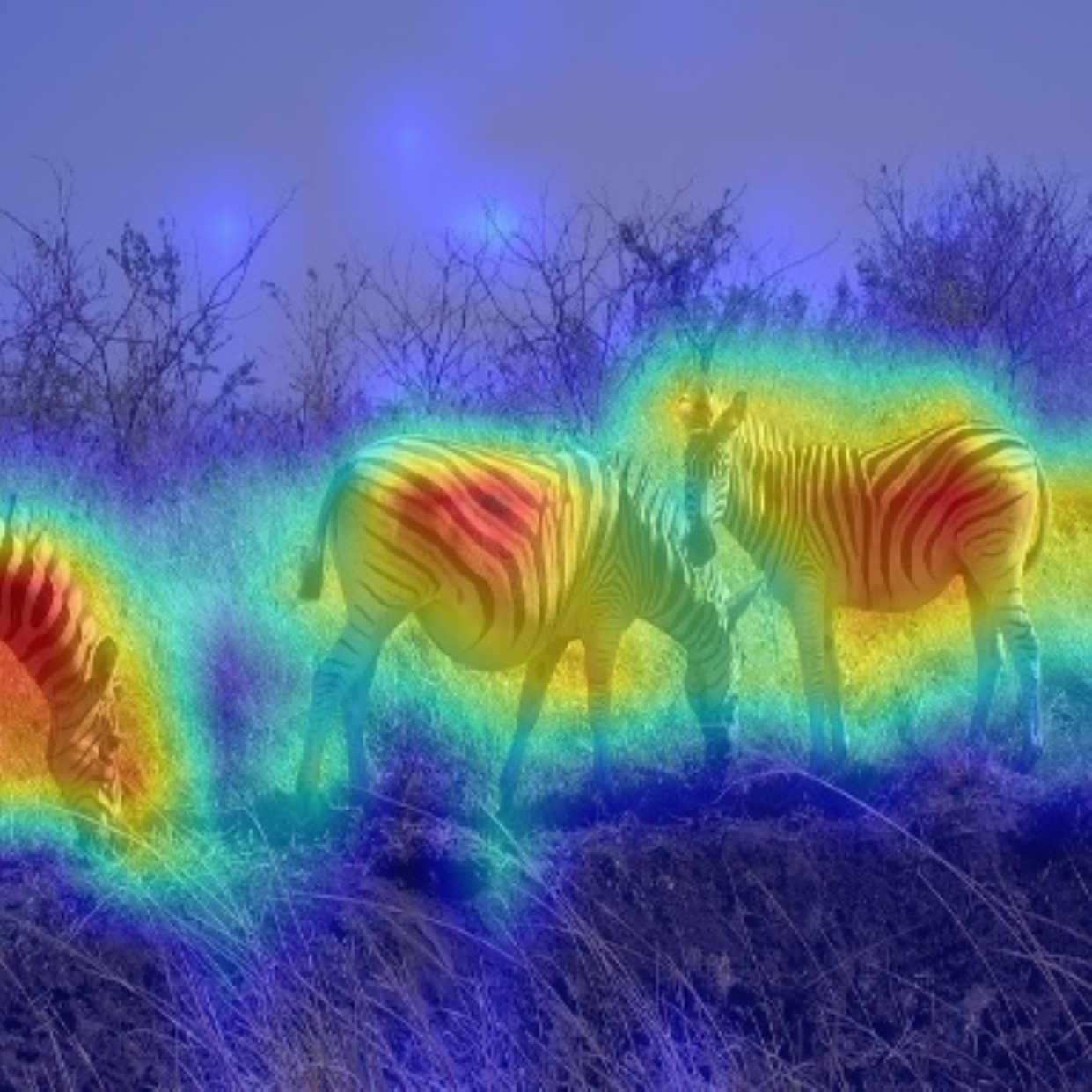} & 
    \includegraphics[width=1.08cm, height=1.08cm]{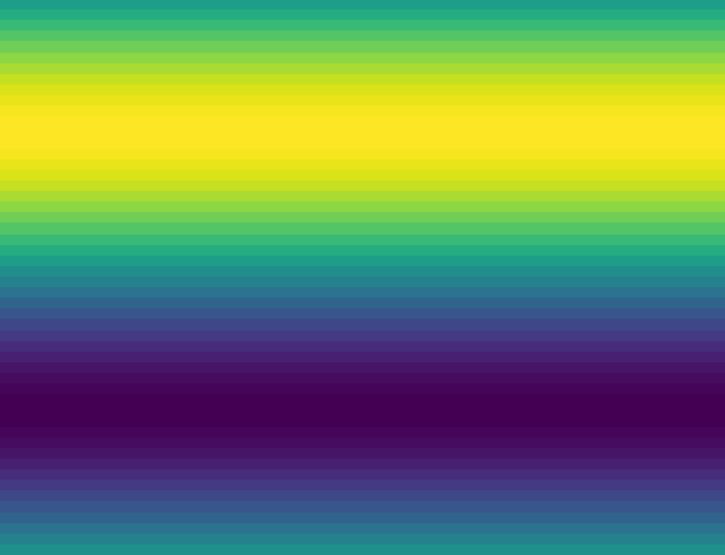} & 
    \includegraphics[width=1.08cm, height=1.08cm]{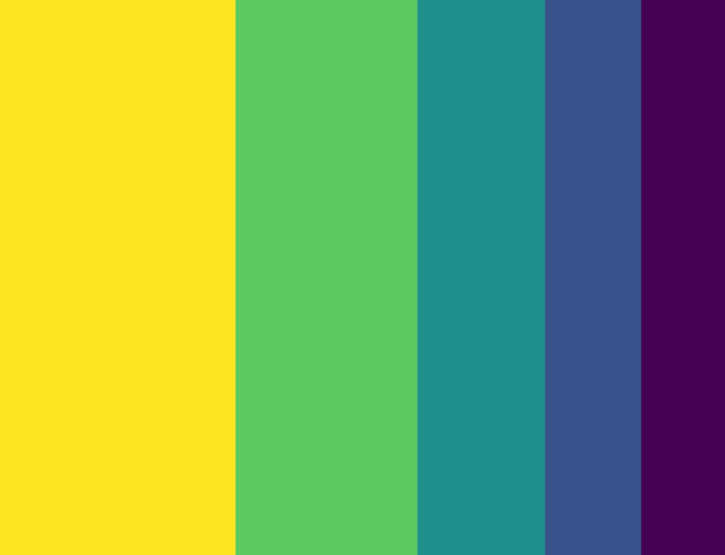} & 
    \includegraphics[width=1.08cm, height=1.08cm]{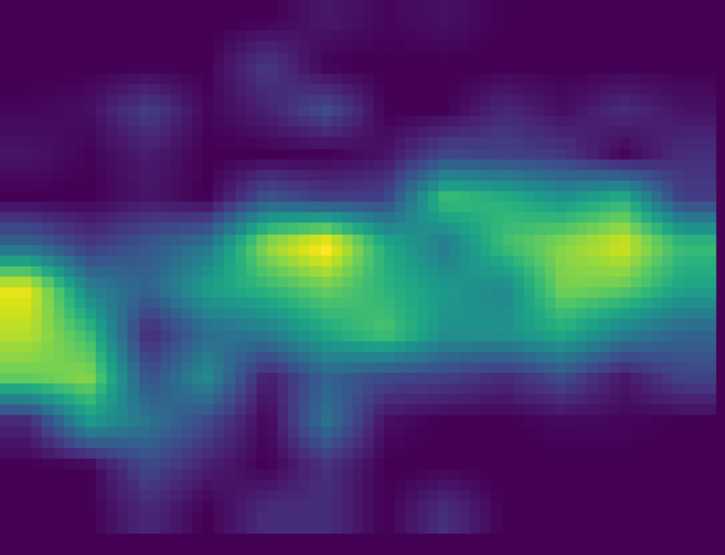} & 
    \includegraphics[width=1.08cm, height=1.08cm]{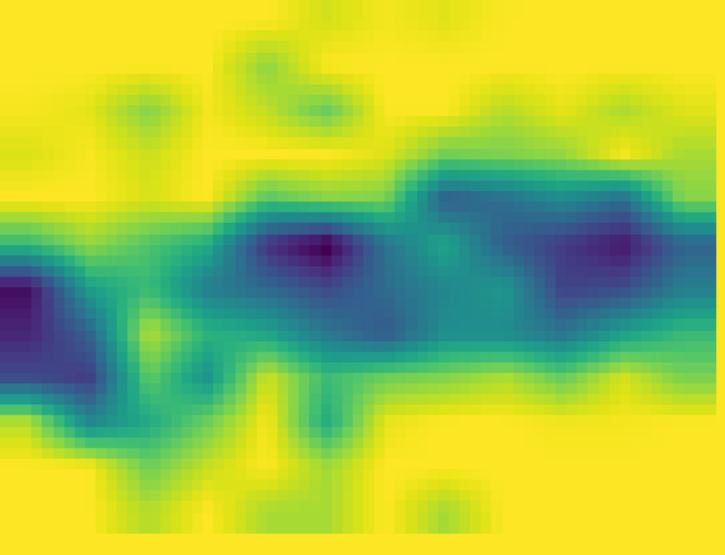} & 
    \includegraphics[width=1.08cm, height=1.08cm]{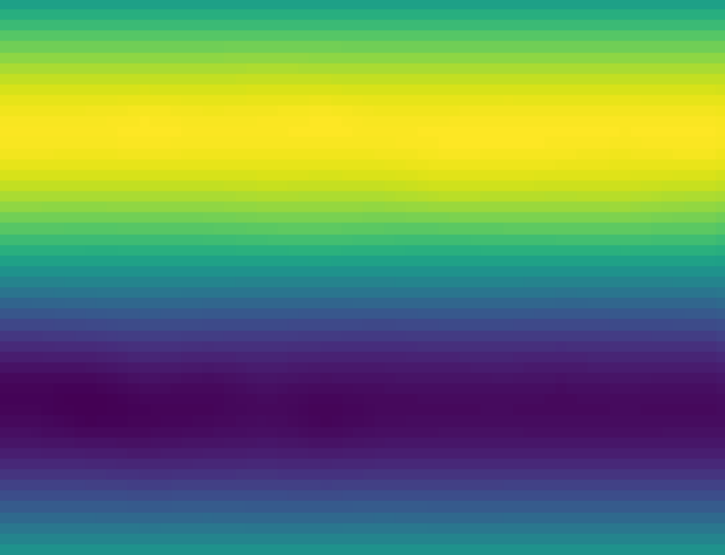} & 
    \includegraphics[width=1.08cm, height=1.08cm]{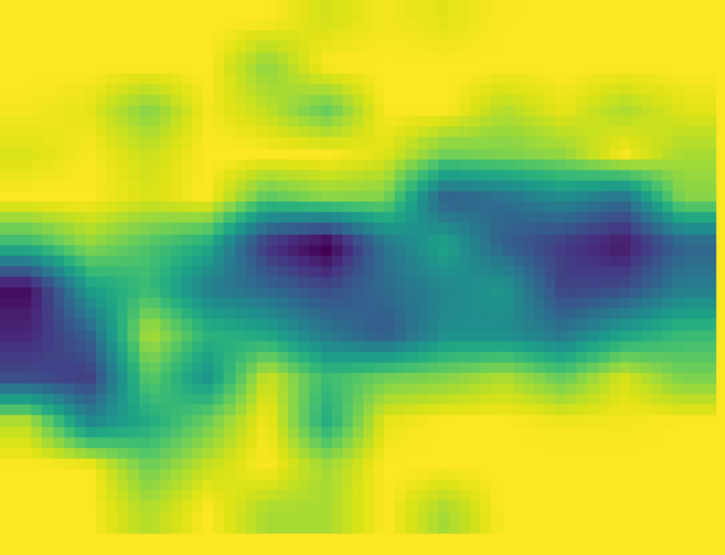} \\ 
    \includegraphics[width=1.08cm, height=1.08cm]{fig/pe/mid/22935.pdf} & 
    \includegraphics[width=1.08cm, height=1.08cm]{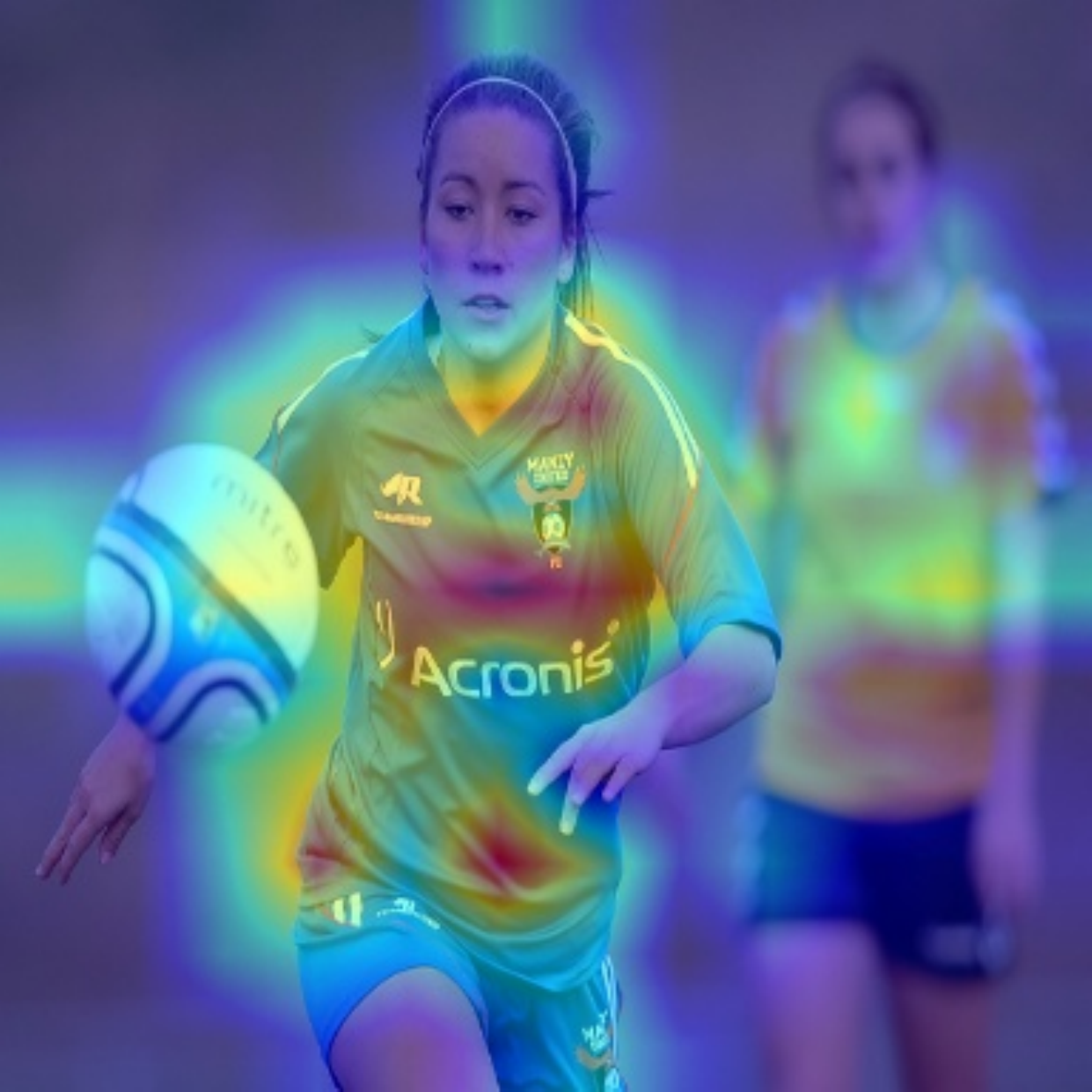} & 
    \includegraphics[width=1.08cm, height=1.08cm]{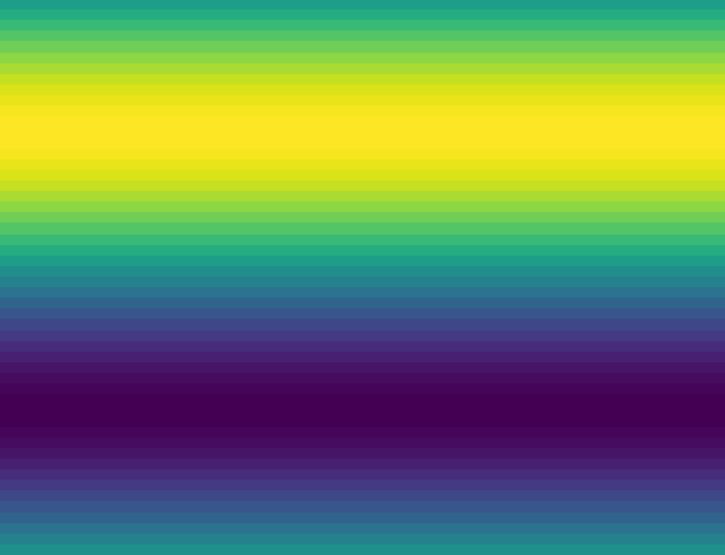} & 
    \includegraphics[width=1.08cm, height=1.08cm]{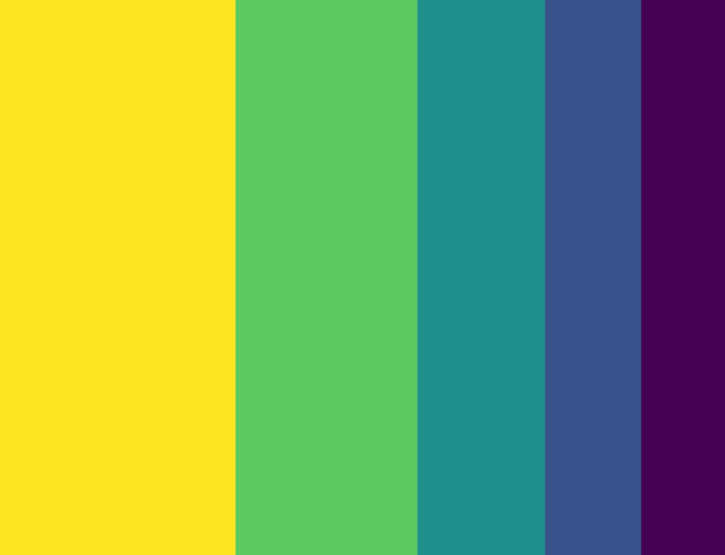} & 
    \includegraphics[width=1.08cm, height=1.08cm]{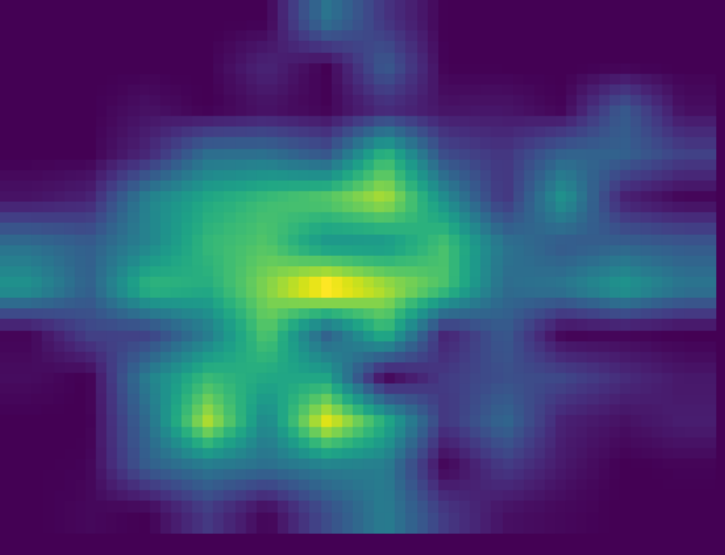} & 
    \includegraphics[width=1.08cm, height=1.08cm]{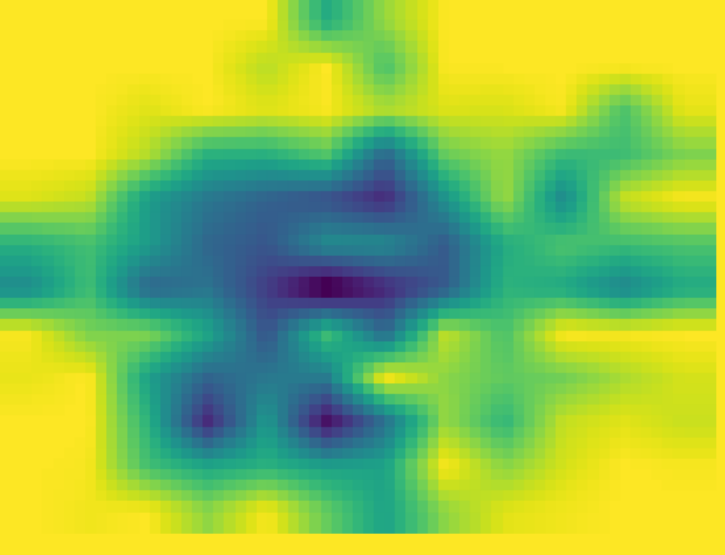} & 
    \includegraphics[width=1.08cm, height=1.08cm]{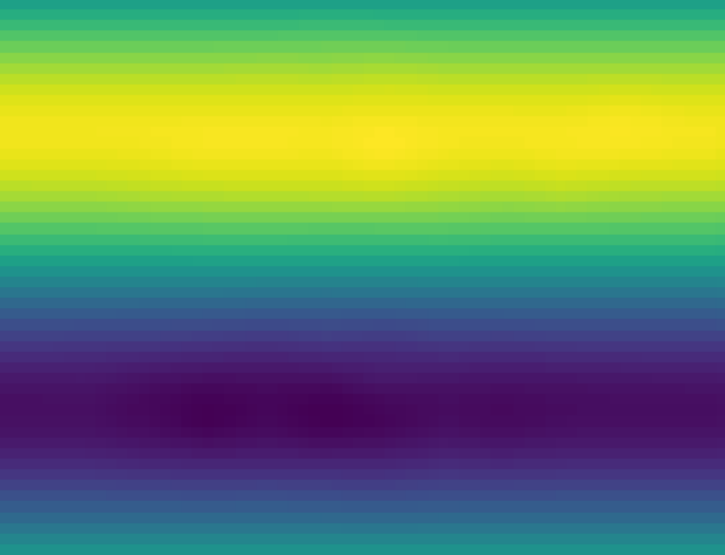} & 
    \includegraphics[width=1.08cm, height=1.08cm]{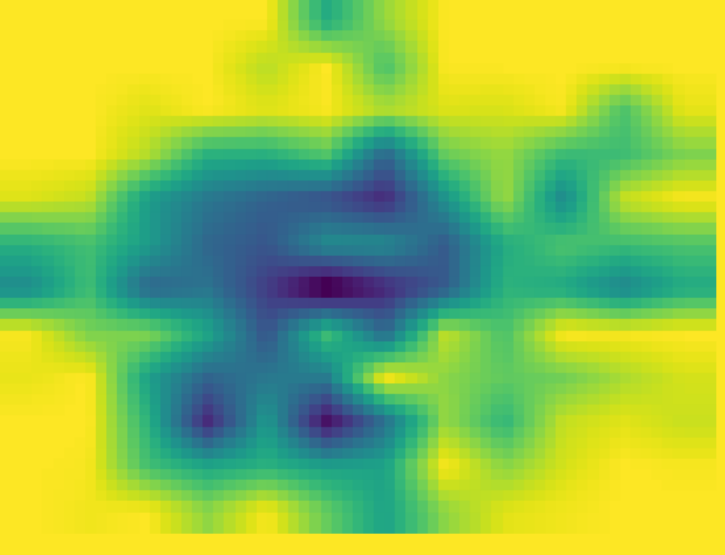} \\ 
    \includegraphics[width=1.08cm, height=1.08cm]{fig/pe/high/9769.pdf} & 
    \includegraphics[width=1.08cm, height=1.08cm]{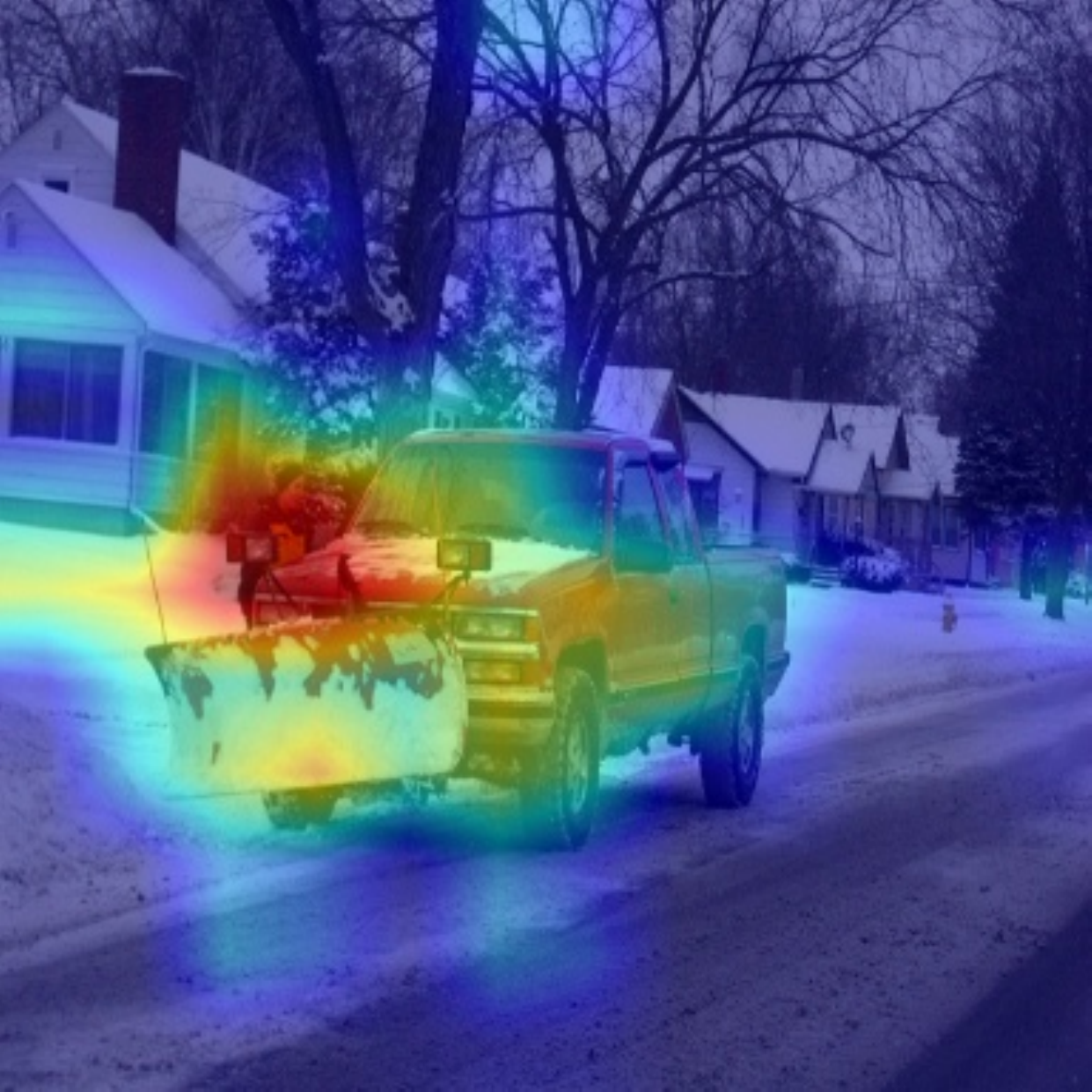} & 
    \includegraphics[width=1.08cm, height=1.08cm]{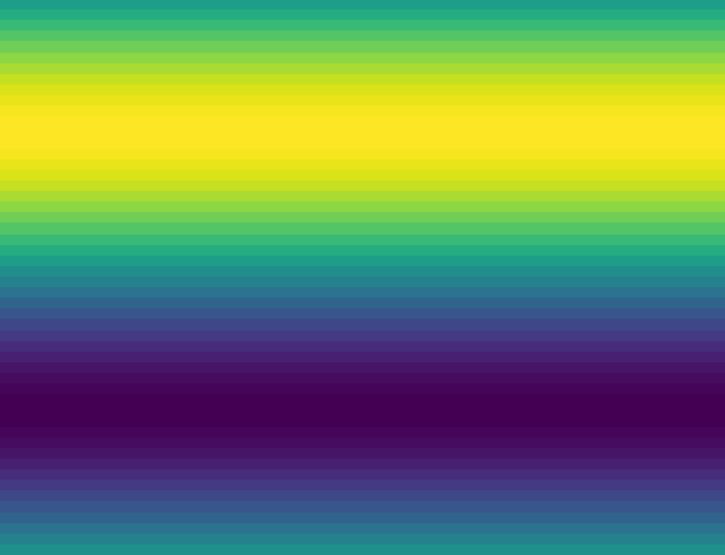} & 
    \includegraphics[width=1.08cm, height=1.08cm]{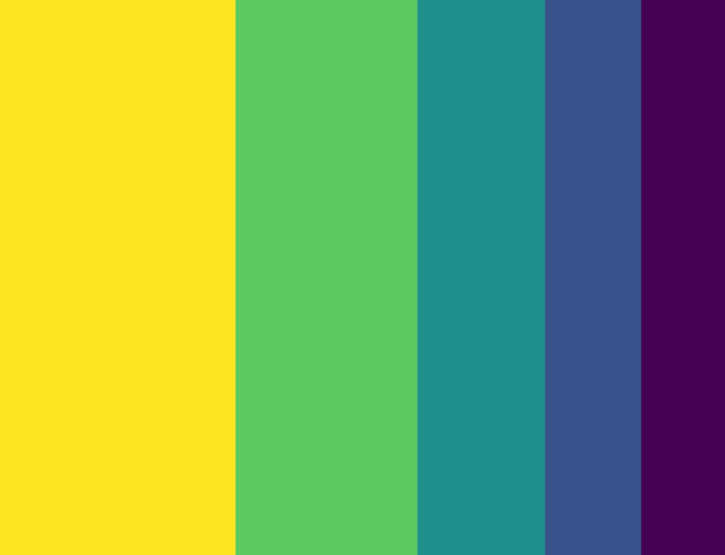} & 
    \includegraphics[width=1.08cm, height=1.08cm]{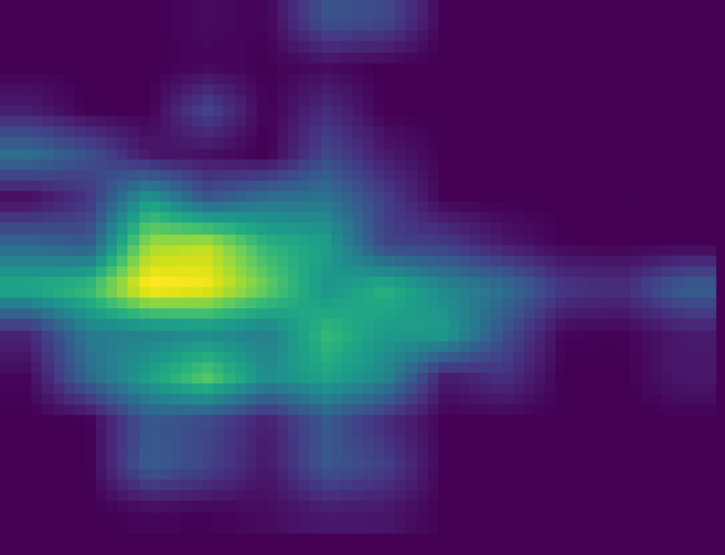} & 
    \includegraphics[width=1.08cm, height=1.08cm]{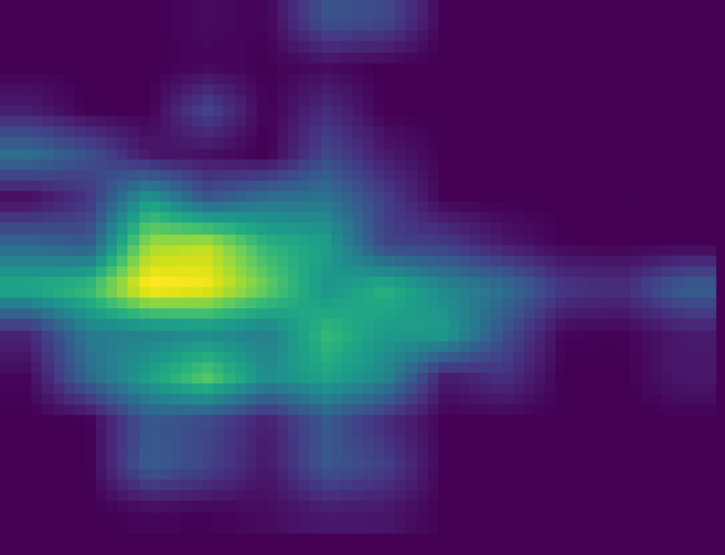} & 
    \includegraphics[width=1.08cm, height=1.08cm]{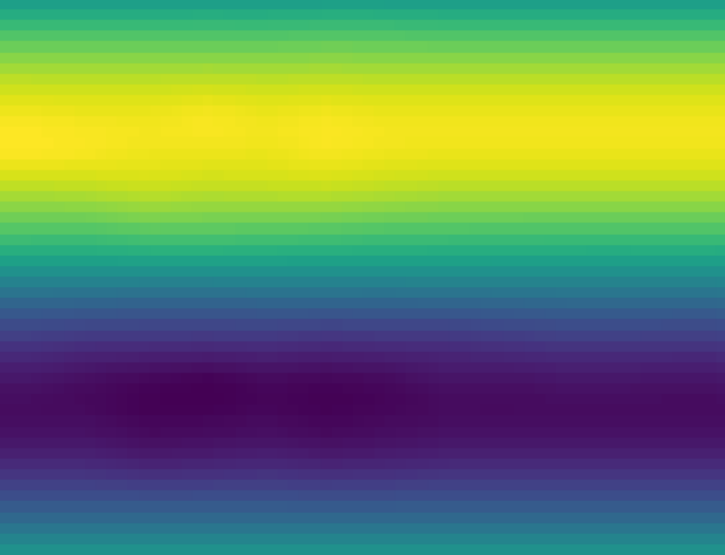} & 
    \includegraphics[width=1.08cm, height=1.08cm]{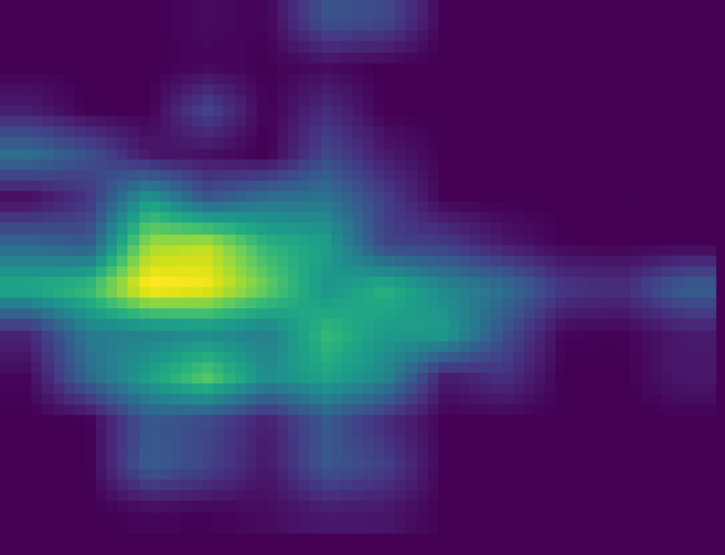} \\ 
    \includegraphics[width=1.08cm, height=1.08cm]{fig/pe/high/169996.pdf} & 
    \includegraphics[width=1.08cm, height=1.08cm]{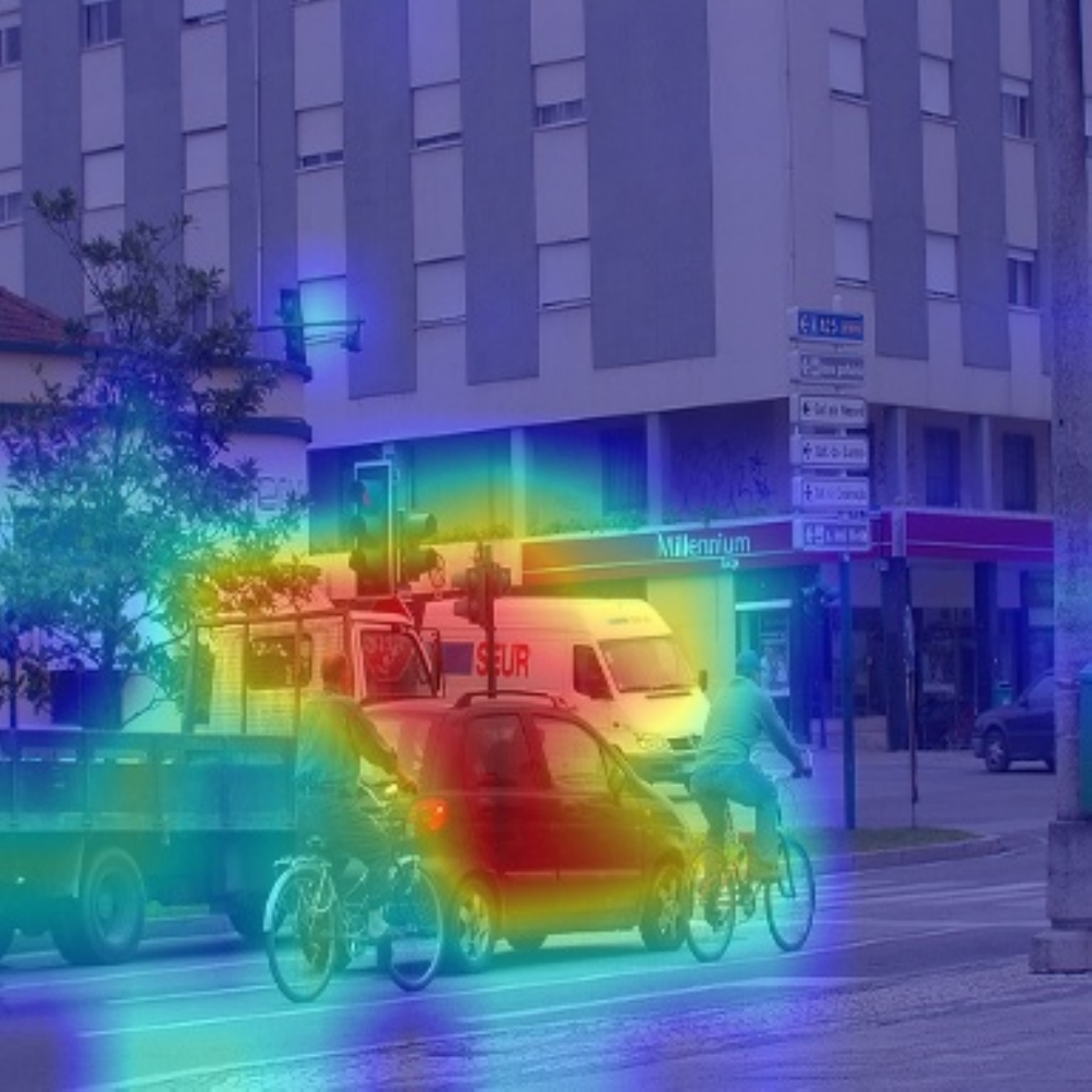} & 
    \includegraphics[width=1.08cm, height=1.08cm]{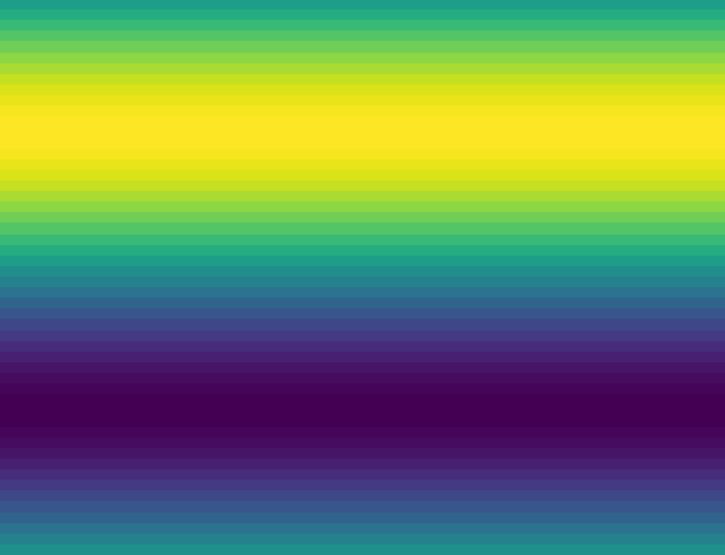} & 
    \includegraphics[width=1.08cm, height=1.08cm]{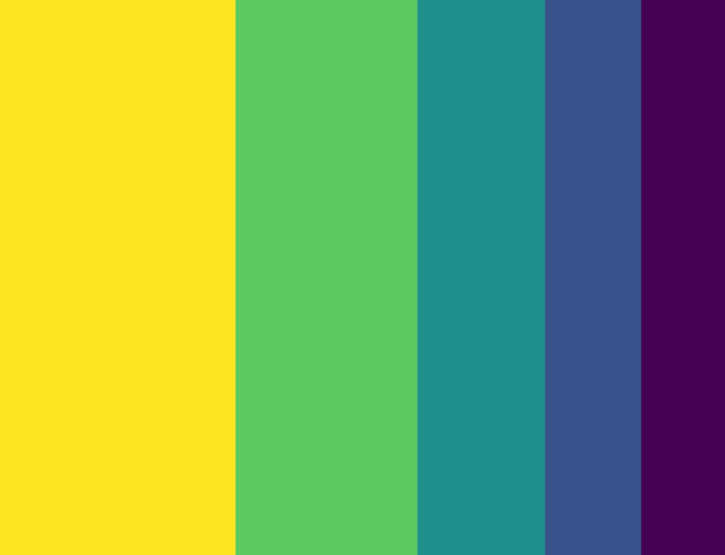} & 
    \includegraphics[width=1.08cm, height=1.08cm]{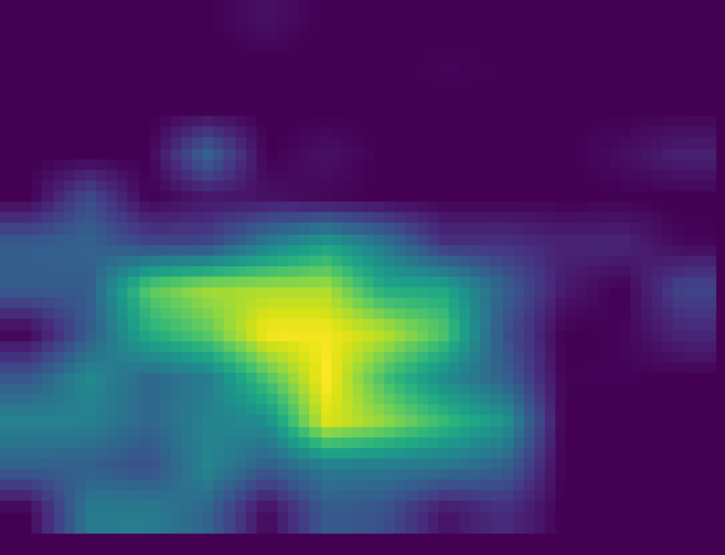} & 
    \includegraphics[width=1.08cm, height=1.08cm]{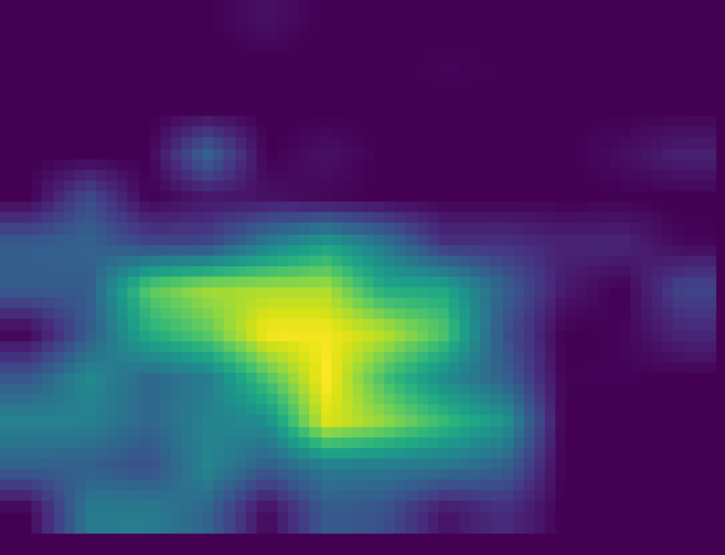} & 
    \includegraphics[width=1.08cm, height=1.08cm]{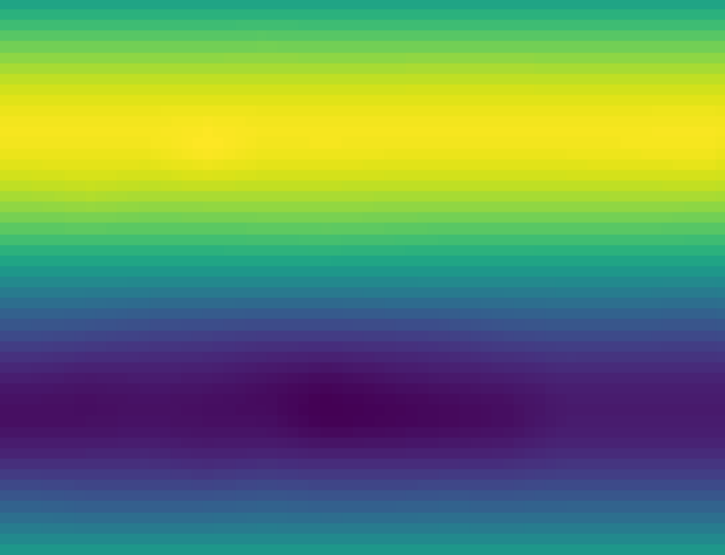} & 
    \includegraphics[width=1.08cm, height=1.08cm]{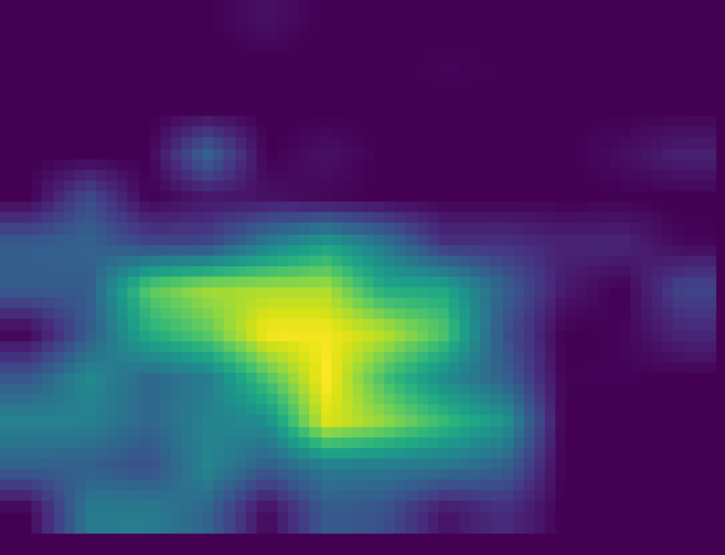} \\[3pt]
    {\scriptsize Image} & {\scriptsize Attention} & {\scriptsize PE-1st} & {\scriptsize PE-last} & {\scriptsize Mod-1st} & {\scriptsize Mod-last} & {\scriptsize PM-1st} & {\scriptsize PM-last}
    \end{tabular}
    \caption{Visualisation of position embedding before and after PEMOLA. From top to bottom, every two rows show low-, medium-, and high-occlusion samples, respectively. PE denotes the original position embedding. Mod denotes the modulation map generated by PEMOLA. PM denotes the modulated position embedding. 1st and last denote the first and last channels. PM-1st and PM-last reveal that the modulated position embedding incorporate occlusion priors, with each channel emphasizing distinct spatial representations.}
   \label{fig:pe_visual}
\end{figure}

\noindent
\textbf{Comparison under Different Occlusion Levels.}
We evaluate the impact of integrating PEMOLA into Mask2Former~\cite{cheng2022masked} and Mask DINO~\cite{li2023mask} on the COCO-OLAC dataset~\cite{wei2025cocoolac} under different occlusion levels in \Cref{table:panoptic_level}. For Mask2Former, PEMOLA yields steady improvements across all occlusion levels, showing notable gains of +1.4 PQ$^{\text{St}}$, +2.4 $\text{AP}_{\text{pan}}^{\text{Th}}$, +3.0 $\text{mIoU}_{\text{pan}}$ under low occlusion and +0.9 PQ under mid occlusion. For Mask DINO, the improvement is more pronounced, showing substantial increases of $+3.3$ PQ, $+4.2$ PQ$^{\text{Th}}$, $+2.0$ PQ$^{\text{St}}$ under low occlusion and $+1.0$ PQ, $+1.5$ PQ$^{\text{Th}}$, $+1.9$ mIoU$_{\text{pan}}$ under high occlusion. This improvement suggests that incorporating occlusion priors helps the models to learn discriminative representations for different occlusion conditions.

\noindent
\textbf{Comparison with SOTA Methods.}
We compare both Mask2Former and Mask DINO with PEMOLA against SOTA methods on the COCO-OLAC dataset in \Cref{table:panoptic_olac}. PEMOLA improves Mask2Former by +0.8 PQ and Mask DINO by +0.8 PQ, with consistent gains in all other metrics. Beyond quantitative improvement, we also provide qualitative visualisation to illustrate why PEMOLA enhances segmentation performance. As shown in~\Cref{fig:pe_visual}, the modulated position embedding effectively captures occlusion patterns, thereby enabling the model to distinguish objects in complex scenes.

We further evaluate Mask2Former with PEMOLA on the Cityscapes-OLAC dataset in \Cref{table:panoptic_city}. PEMOLA brings a +0.8 PQ improvement over Mask2Former, along with additional gains of +1.4 $\text{PQ}^{\text{Th}}$, +3.3 $\text{AP}_{\text{pan}}^{\text{Th}}$, and +1.3 $\text{mIoU}_{\text{pan}}$. This confirms that PEMOLA maintains robustness across datasets with distinct data distributions.

\begin{table}[!t]
\caption{Ablation study on the proposed occlusion priors.}
\centering
\begin{adjustbox}{width={0.48\textwidth}, keepaspectratio}
\begin{tabular}{@{\hspace{4pt}}l|c c c|c c@{\hspace{4pt}}}
\toprule
 & \text{PQ} & $\text{PQ}^{\text{Th}}$ & $\text{PQ}^{\text{St}}$ & $\text{AP}_{\text{pan}}^{\text{Th}}$ & $\text{mIoU}_{\text{pan}}$ \\ 
\midrule
Ours & \textbf{41.5} & \textbf{45.2} & \textbf{35.9} & \textbf{30.4} & \textbf{54.8} \\
\midrule
- occlusion label embedding & 41.0 & 44.8 & 35.3 & 30.4 & 54.7 \\
- occlusion-level attention & 40.9 & 44.6 & 35.3 & 30.2 & 54.2 \\
\midrule
- remove all the above & 40.7 & 44.5 & 35.0 & 30.0 & 54.2 \\
\bottomrule
\end{tabular}
\end{adjustbox}
\label{table:pemola_priors}
\end{table}

\begin{table}[!t]
\caption{Ablation study on the Grad-CAM smoothing.}
\centering
\begin{adjustbox}{width={0.42\textwidth}, keepaspectratio}
\begin{tabular}{@{\hspace{4pt}}l|c c c|c c@{\hspace{4pt}}}
\toprule
 & \text{PQ} & $\text{PQ}^{\text{Th}}$ & $\text{PQ}^{\text{St}}$ & $\text{AP}_{\text{pan}}^{\text{Th}}$ & $\text{mIoU}_{\text{pan}}$ \\ 
\midrule
Base & 40.7 & 44.5 & 35.0 & 30.0 & 54.2 \\
\midrule
w/o smooth & 41.0 & 44.7 & 35.4 & 30.3 & 54.4 \\
w smooth (ours) & \textbf{41.5} & \textbf{45.2} & \textbf{35.9} & \textbf{30.4} & \textbf{54.8} \\
\bottomrule
\end{tabular}
\end{adjustbox}
\label{table:pemola_cam}
\end{table}

\begin{table}[!t]
\caption{Ablation study on the insertion location of PEMOLA.}
\centering
\begin{adjustbox}{width={0.433\textwidth}, keepaspectratio}
\begin{tabular}{@{\hspace{4pt}}l|c c c|c c@{\hspace{4pt}}}
\toprule
 & \text{PQ} & $\text{PQ}^{\text{Th}}$ & $\text{PQ}^{\text{St}}$ & $\text{AP}_{\text{pan}}^{\text{Th}}$ & $\text{mIoU}_{\text{pan}}$ \\ 
\midrule
Base & 40.7 & 44.5 & 35.0 & 30.0 & 54.2 \\
\midrule
transformer decoder & 40.7 & 44.2 & 35.2 & 30.3 & 54.2 \\
pixel decoder (ours) & \textbf{41.5} & \textbf{45.2} & \textbf{35.9} & \textbf{30.4} & \textbf{54.8} \\
\bottomrule
\end{tabular}
\end{adjustbox}
\label{table:pemola_location}
\end{table}

\subsection{Ablation Study}
We conduct a series of ablation studies to evaluate the contribution of each component in PEMOLA using Mask2Former on the COCO-OLAC dataset.

\noindent
\textbf{Occlusion Priors.}
We evaluate the effectiveness of the proposed occlusion priors in \Cref{table:pemola_priors}. Removing the occlusion label embedding or the occlusion-level attention leads to a PQ drop of 0.5 and 0.6, respectively, indicating that both components contribute to the overall improvement.

\noindent
\textbf{Grad-CAM Smoothing.}
Grad-CAM includes two smoothing strategies, \textit{eigen smoothing} and \textit{augmentation smoothing}, which help reduce gradient noise and stabilise the attention maps. As shown in \Cref{table:pemola_cam}, applying these methods marginally improves PQ from 41.0 to 41.5.

\noindent
\textbf{Insertion Location.}
\label{subsec: insert location}
We analyse the effect of the insertion location of PEMOLA in \Cref{table:pemola_location}. The results show that PEMOLA is effective only when applied to the pixel decoder. We hypothesise that the difference arises 
because the occlusion attention is consistent with the feature representations in the pixel decoder, where the spatial layout and occlusion patterns remain closely aligned. In contrast, the transformer decoder operates on task-orientated embeddings with different feature distributions, making the modulation less compatible.

\section*{Acknowledgments}

Calculations were performed using the University of Warwick's SCRPT Sulis Tier 2 HPC platform, funded by EPSRC Grant EP/T022108/1 and the HPC Midlands+ consortium.

\section{CONCLUSIONS}
In this paper, we propose PEMOLA, a novel occlusion-aware module that enhances transformer-based panoptic segmentation by integrating occlusion priors into position embedding. PEMOLA jointly leverages spatial occlusion-level attention and channel-wise occlusion label embedding to adaptively modulate positional representations under varying occlusion conditions. To support cross-dataset evaluation, we also introduce Cityscapes-OLAC as an occlusion-annotated extension of Cityscapes. Extensive experiments on the COCO-OLAC and Cityscapes-OLAC datasets demonstrate that PEMOLA consistently improves segmentation quality with minimal computational overhead. The results highlight that incorporating occlusion priors into the representation learning is an effective way to improve reliability in complex scenes. Future work will extend this framework to video panoptic segmentation and explore dynamic occlusion reasoning in temporal contexts.

\bibliographystyle{IEEEbib}
\bibliography{icme2025references}

\end{document}